\documentclass[11pt]{article}

\usepackage[final]{acl}

\usepackage{times}
\usepackage{latexsym}

\usepackage[T1]{fontenc}

\usepackage[utf8]{inputenc}

\usepackage{microtype}

\usepackage{inconsolata}

\usepackage{graphicx}

\usepackage{listings}     %

\usepackage[utf8]{inputenc} %
\usepackage[T1]{fontenc}    %
\usepackage{hyperref}       %
\usepackage{url}            %
\usepackage{booktabs}       %
\usepackage{amsfonts}       %
\usepackage{nicefrac}       %
\usepackage{microtype}      %
\usepackage[dvipsnames]{xcolor}        %
\usepackage{makecell}
\usepackage{tabularx}
\usepackage{wrapfig}
\usepackage{inconsolata}
\usepackage{multirow}
\usepackage{graphicx}
\usepackage{enumitem}
\usepackage{tcolorbox}
\usepackage{color-edits}
\addauthor{gn}{magenta}
\addauthor{ysu}{cyan}
\definecolor{lightgreen}{RGB}{145, 204, 117}
\definecolor{lightyellow}{RGB}{250, 200, 88}
\definecolor{lightred}{RGB}{238, 102, 102}

\newtcolorbox{promptbox}[2][Prompt]{
colback=black!5!white,
arc=5pt, 
boxrule=0.5pt,
fonttitle=\bfseries,
title=#1, 
before upper={\small}, fontupper=\fontfamily{ptm}\selectfont,
colframe=#2,
}

\usepackage{enumitem}
\usepackage{tabularx} 
\usepackage{float}

\title{SciVQR: A Multidisciplinary Multimodal Benchmark for Advanced Scientific Reasoning Evaluation}

\author{
 \textbf{Longteng Guo\textsuperscript{1,2,$\ast$}},
 \textbf{Xuanxu Lin\textsuperscript{1,2,$\ast$}},
 \textbf{Dongze Hao\textsuperscript{3,$\ast$}},
 \textbf{Tongtian Yue\textsuperscript{1,2}},
\\
 \textbf{Pengkang Huo\textsuperscript{1,2}},
 \textbf{Jiatong Ma\textsuperscript{1,2}},
 \textbf{Yuchen Liu\textsuperscript{1,2}},
 \textbf{Jing Liu\textsuperscript{1,2,$\dagger$}}
\\
\\
 \textsuperscript{1}Institute of Automation, Chinese Academy of Sciences (CASIA)\\
 \textsuperscript{2}School of Artificial Intelligence, University of Chinese Academy of Sciences (UCAS)\\
 \textsuperscript{3}OPPO AI Center
\\
\texttt{\{longteng.guo, jliu\}@nlpr.ia.ac.cn} \\
\texttt{linxuanxu2025@ia.ac.cn, haodongze@oppo.com}
}

\newcommand\blfootnote[1]{%
  \begingroup
  \renewcommand\thefootnote{}\footnote{#1}%
  \addtocounter{footnote}{-1}%
  \endgroup
}

\begin{document}
\maketitle
\blfootnote{$^\ast$Equal contribution.}
\blfootnote{$^\dagger$Corresponding author.}

\begin{abstract}
Scientific reasoning is a key aspect of human intelligence, requiring the integration of multimodal inputs, domain expertise, and multi-step inference across various subjects. Existing benchmarks for multimodal large language models (MLLMs) often fail to capture the complexity and traceability of reasoning processes necessary for rigorous evaluation. To fill this gap, we introduce SciVQR, a multimodal benchmark covering 54 subfields in mathematics, physics, chemistry, geography, astronomy, and biology. SciVQR includes domain-specific visuals, such as equations, charts, and diagrams, and challenges models to combine visual comprehension with reasoning. The tasks range from basic factual recall to complex, multi-step inferences, with 46\% including expert-authored solutions. SciVQR not only evaluates final answers but also examines the reasoning process, providing insights into how models reach their conclusions. Our evaluation of leading MLLMs, including both proprietary and open-source models, reveals significant limitations in handling complex multimodal reasoning tasks, underscoring the need for improved multi-step reasoning and better integration of interdisciplinary knowledge in advancing MLLMs toward true scientific intelligence. The dataset and evaluation code are publicly available at \url{https://github.com/CASIA-IVA-Lab/SciVQR}.
\end{abstract}

\section{Introduction}
The rapid advancement of large-scale models, particularly multimodal large language models (MLLMs) \citep{li2024llava, zhu2023minigpt, trinh2024solving}, has brought us closer to achieving human-level intelligence, capable of processing and reasoning across diverse modalities such as text and images.
Among human cognitive abilities, scientific reasoning is particularly crucial: analyzing multimodal cues, performing multi-step inference, integrating domain knowledge, and drawing systematic conclusions across diverse scientific fields.
Despite advances in natural language understanding, current models still struggle with tasks requiring deep, structured reasoning grounded in multimodal scientific content.
To rigorously assess progress in this direction, it is essential to benchmark MLLMs on their ability to perform scientific reasoning involving diverse subjects, multimodal inputs, and challenging multi-step reasoning.

We highlight three principles necessary for such evaluation:

\textbf{(1) Disciplinary coverage:} Scientific reasoning spans far beyond a few canonical subjects like physics or mathematics. A truly comprehensive benchmark should cover a broad range of scientific domains, reflecting the diversity of real-world knowledge.

\textbf{(2) Difficulty diversity:} A well-designed benchmark should cover a wide spectrum of difficulty levels, ranging from basic factual or conceptual reasoning to college- and graduate-level, multi-hop inference. This enables more nuanced evaluation across different reasoning depths and helps prevent performance saturation at either end of the spectrum.

\textbf{(3) Solution traceability:} Evaluating only the final answer is insufficient for understanding a model's reasoning capabilities. Benchmarks should include key intermediate solution steps to enable fine-grained analysis of reasoning fidelity and structure.

While existing benchmarks address parts of this challenge, none cover all dimensions. For example, MMMU \citep{Yue_2024_CVPR} spans multiple subjects but lacks difficulty diversity and reasoning traces. ScienceQA \citep{lu2022learn} includes solution steps but is limited to K-12 content with shallow reasoning. OmniMath \citep{gao2024omni} targets advanced mathematical reasoning but is narrow in scope and monolingual.

As a result, there remains a clear gap in benchmarking the emerging scientific capabilities of next-generation MLLMs, which requires disciplinary breadth, reasoning depth, and cross-linguistic generalization.

\begin{figure*}[!h]
    \centering
    \includegraphics[width=\textwidth]{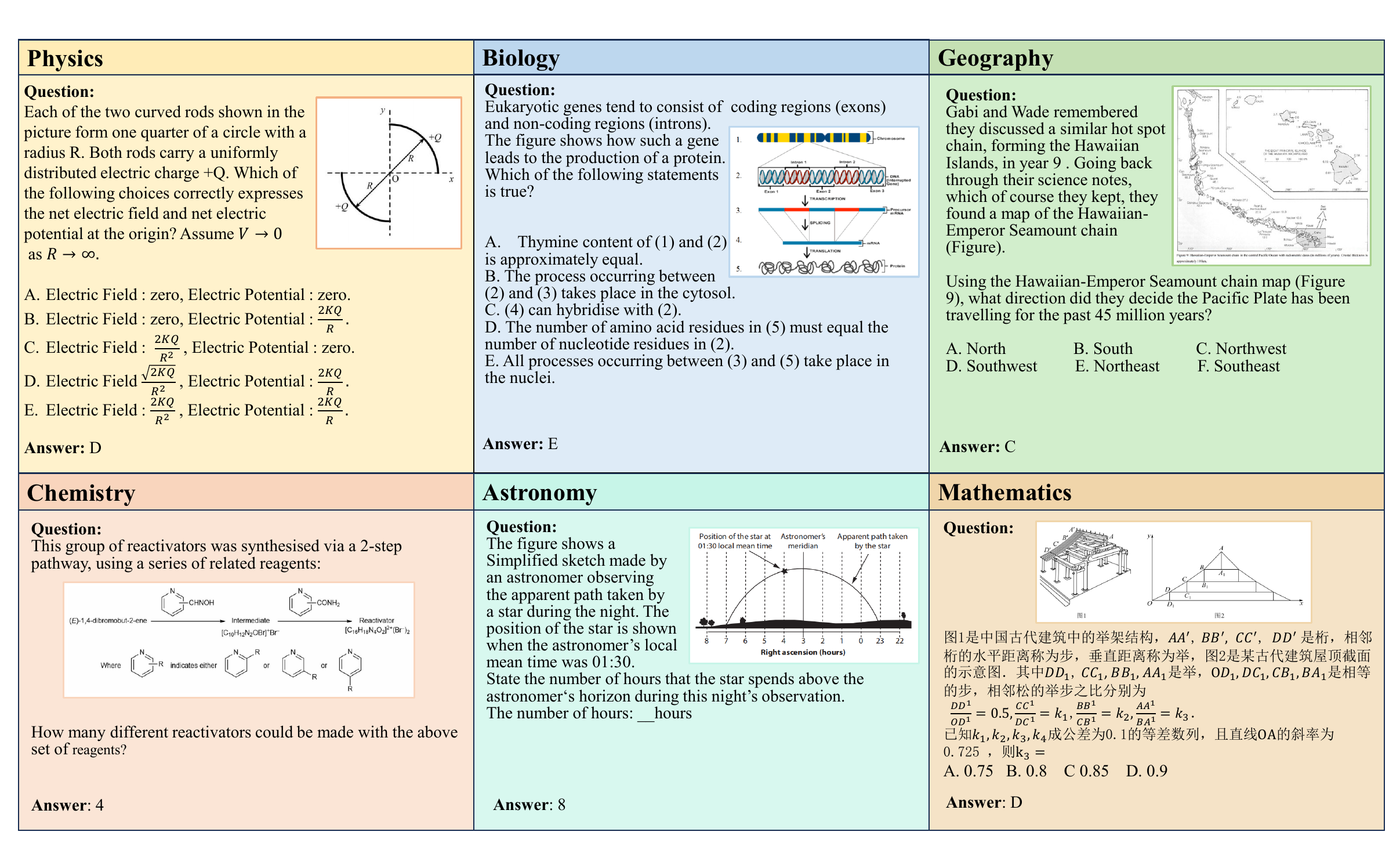}
    \caption{Sampled SciVQR examples from each subject. To solve the questions, the models need strong multimodal reasoning capabilities.}
    \label{fig:examples}
\end{figure*}

To bridge this gap, we introduce SciVQR, a comprehensive multimodal benchmark for scientific reasoning in MLLMs. Covering 54 subfields across 6 core scientific domains (mathematics, physics, chemistry, geography, astronomy, and biology), SciVQR ensures broad disciplinary representation. Figure~\ref{fig:examples} illustrates sampled questions from each subject.

The questions are sourced from high school, college, and graduate-level exams, quizzes, textbooks, and competitions, and span a wide range of difficulty levels (easy, medium, and hard). Many require advanced, multi-step reasoning, such as solving differential equations or interpreting complex molecular structures. Notably, 46\% of the questions are accompanied by detailed, expert-authored solution traces to support precise reasoning evaluation. SciVQR also incorporates diverse scientific image formats, such as equations, charts, chemical structures, geological diagrams, and biological illustrations. %
It is developed through a rigorous, multi-stage quality assurance process involving both human experts and LLM assistance.
SciVQR challenges models to integrate fine-grained visual understanding, deep subject knowledge, and sophisticated reasoning, thereby setting a new standard for scientific multimodal evaluation in MLLMs.

We evaluate a series of open-source and advanced proprietary MLLMs, including GPT-4o \citep{hurst2024gpt} and o1 \citep{jaech2024openai}, on SciVQR. Our key findings are summarized as follows:
\begin{itemize}[left=0pt, labelsep=1em]
\item \textbf{Narrowing gap between proprietary and open-source models:} While there remains a substantial performance gap between proprietary models and open-source models, leading open-source models like Qwen2.5-VL-72B \citep{bai2025qwen2} (61.8\%) and the reasoning-optimized MiMo-VL-7B \citep{xiaomi2025mimo} (69.2\%) have started to close the gap, even surpassing non-reasoning proprietary models such as GPT-4o \citep{hurst2024gpt} (60.7\%).

\item \textbf{Disciplinary performance differences:} Models perform significantly worse in mathematics and physics due to the greater emphasis on quantitative analysis and reasoning. In contrast, subjects like biology and geography, which rely more on textual understanding and knowledge, see relatively better performance.

\item \textbf{Reasoning optimization leads to better performance:} Reasoning-optimized models such as o1 \citep{jaech2024openai} (77.2\%), o4-mini \citep{openai2024systemcard} (79.3\%), and MiMo-VL \citep{xiaomi2025mimo} (69.2\%) substantially outperform non-reasoning models like GPT-4o \citep{hurst2024gpt} (60.7\%), with marked improvements across all subjects. Notably, mathematics and physics scores experience the greatest boost, underscoring the importance of reasoning capabilities.

\item \textbf{Reasoning optimization improves reasoning traces:} Our CoT evaluations reveal that reasoning-optimized models tend to generate more faithful, complete, and structured reasoning traces. This suggests that optimizing for step-by-step reasoning, rather than relying solely on instruction tuning, is crucial for improving scientific inference performance.

\end{itemize}

Our findings emphasize that achieving true multimodal scientific reasoning requires deeper multi-step inference and the integration of diverse knowledge across subjects.

\section{Related Work}

\subsection{Multimodal Large Language Models}

The development of MLLMs \cite{liu2023visual,li2024llava,li2023blip,wang2024qwen2,bai2025qwen2,liu2025vrope,liu2024improved,wu2024deepseek} builds on advances in multimodal learning and the success of LLMs \cite{achiam2023gpt,yang2024qwen2,liu2024deepseek}. Early vision-language frameworks established the foundation for modern systems, which now combine strong vision encoders with LLMs to handle diverse visual inputs. Open-source initiatives and scalable training, including visual instruction tuning \cite{liu2023visual} and synthetic data generation \cite{zhang2024video,maaz2023video}, have improved adaptability and generalization. While closed-source models like GPT-4o \cite{hurst2024gpt} show strong visual reasoning, open-source models such as LLaVA \cite{liu2023visual}, MiniGPT-4 \cite{zhu2023minigpt}, and Qwen2-VL \cite{wang2024qwen2} advance progress by refining architectures, expanding datasets, and broadening applications.

Reasoning capabilities of MLLMs have evolved alongside text-based LLMs \cite{wei2022chain,guo2025deepseek}. Though once criticized as mere token predictors \cite{huang2023towards}, LLMs now demonstrate skill in logic \cite{xin2024deepseek}, problem-solving \cite{rein2024gpqa}, and code generation \cite{hui2024qwen2}. This has spurred multimodal reasoning research, with methods like Visual-RFT \cite{liu2025visual}, VLM-R1 \cite{shen2025vlm}, and Vision-R1 \cite{huang2025vision}, emphasizing perception-reasoning integration for complex visual tasks.

\subsection{Multimodal Reasoning Evaluation}

Evaluation of multimodal systems has progressed from early benchmarks such as VQA \cite{antol2015vqa} and MSCOCO \cite{lin2014microsoft}, which tested perception and alignment, to more comprehensive ones. Recent benchmarks like POPE \cite{li2023evaluating}, LAMM \cite{yin2023lamm}, MMBench \cite{xu2023mmbench}, and SEED \cite{li2024seed} expand coverage to hallucination detection, cross-modal reasoning, and real-world simulation. They assess capabilities from low-level recognition to high-level contextual understanding, reflecting increasing model complexity.

For reasoning-specific evaluation, targeted benchmarks focus on scientific and mathematical reasoning. MathVista \cite{lu2024mathvista} and OlympiadBench \cite{he2024olympiadbench} integrate diagrams, equations, and domain knowledge. MathVerse \cite{zhang2024mathverse} removes textual redundancy in visual problems to test pure graphical reasoning. MMMU-Pro \cite{yue2024mmmu} further introduces filtered protocols, vision-only settings, and adversarial questions to avoid text shortcuts. These works mark a shift toward inseparable modality fusion, where visual analysis and iterative reasoning are essential.

Despite these advances, none provide a comprehensive evaluation of scientific reasoning. Many focus on perception or commonsense reasoning, while domain-specific ones face limitations. Some target only K-12 content with shallow reasoning (e.g., ScienceQA \cite{lu2022learn}), while others (e.g., EMMA \cite{hao2025can}) restrict themselves to mathematics, physics, and chemistry, missing broader scientific knowledge. Most benchmarks also provide only final answers, without step-by-step solutions needed to assess reasoning fidelity.

\section{The SciVQR Benchmark}
\subsection{Overview of SciVQR}

We present SciVQR, a carefully curated benchmark developed to assess the multimodal scientific reasoning capabilities of foundation models across a diverse set of visually grounded scientific tasks. As shown in Figure~\ref{fig:subjects}, SciVQR covers 6 scientific subjects, including mathematics, physics, chemistry, astronomy, geography, and biology, and each subject contains various specialized subfields. The questions in our benchmark were manually collected from different academic competitions, college and graduate-level exam sets, and college textbooks. We categorize all questions into easy, medium, and hard levels with reference to their original source materials.

\subsection{Data Curation Process}
\paragraph{Data Collection:} Our benchmark is built in two stages. First, we collect competitions, graduate exams, college exams, and textbooks across multiple subjects, prioritizing resources rich in images to ensure sufficient visual questions. This yields 15 scientific competitions, 9 college entrance exams, and 6 college or graduate-level exams and textbooks from different countries. In the second stage, we recruit over 20 postgraduates to annotate the questions, following copyright and licensing rules to exclude restricted sources. All annotators receive fair compensation based on the local competitive hourly rate. To mitigate data contamination and language priors, we select questions where visual information is indispensable rather than auxiliary. Annotators capture screenshots of visual questions, then extract text using the Mathpix OCR API. All images are standardized to PNG format, and since each image naturally aligns with one question, we obtain one-to-one image-question pairs. This process produces a diverse pool of 5k questions.

\paragraph{Data Quality Control:} To ensure the quality of the collected data, we use a web-based tool to filter the annotated questions. In the first stage, we exclude samples with low-resolution or incorrect images, and verify the correctness of the textual content, ensuring that all examples conform to the required quality standards. In the second stage, we use lexical overlap and Levenshtein distance to identify and remove duplicate problems. After these steps, we obtain the final 3,254 science-VQA instances for our benchmark. In the third stage, we categorize the problems of different subjects into different subfields. We first provide a predefined list of possible subfields to GPT-4V \citep{openai2023gpt4v}, prompting the model to assign an initial subfield category to each question. We then carefully review and correct these preliminary labels. Subsequently, we assign a difficulty level to each question, guided by its original source material. Across all these stages, we perform iterative manual verification and consistency checks to ensure the reliability of the final annotations.

\subsection{Data Analysis}

\begin{table}[htbp]
  \centering
  \small
  \begin{tabular}{lr}
    \toprule
    \textbf{Statistic} & \textbf{Number}\\
    \midrule
    Total questions & 3,254\\
    - Multiple-choice questions & 2,545 (78.2\%)\\
    - Free-form questions & 709 (21.8\%)\\
    - Questions with an explanation &1490 (45.8\%)\\
    \midrule
    English questions &2661 (81.8\%)\\
    Chinese questions &593 (18.2\%)\\
    \midrule
    Subjects & 6 \\
    Subfields &  54\\
    Difficulty levels & 3 \\
    \midrule
    Maximum question length & 495\\ 
    Average question length & 77.2\\
    \midrule
    Maximum explanation length & 782\\ 
    Average explanation length & 102.4\\
  \bottomrule
\end{tabular}
\caption{Main statistics in SciVQR.}
\label{tab:dataset}
\vspace{-10pt}
\end{table}

\paragraph{Key Statistics.}
Table~\ref{tab:dataset} summarizes the main statistics of SciVQR. The dataset contains 3,254 questions, balancing complexity and practicality with both structured and unstructured formats. Specifically, 78.2\% of the questions are multiple-choice, offering a controlled evaluation setting, while the remaining 21.8\% are free-form, encouraging open-ended reasoning and expression. Notably, 45.8\% of all questions include explanations, offering insights into reasoning processes that can support interpretability and educational analysis.

\begin{figure*}[htbp]
    \centering
    \includegraphics[width=\linewidth]{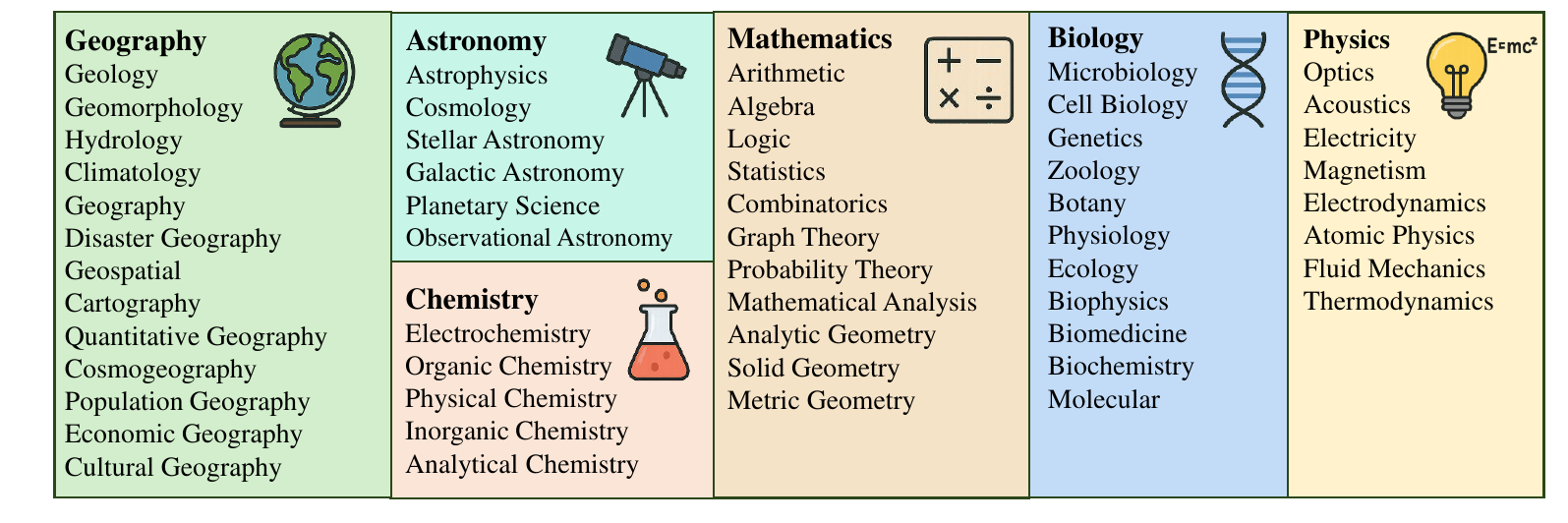}
    \caption{Domain diversity in SciVQR. Each color corresponds to one subject: mathematics, physics, chemistry, biology, astronomy, geography. Each subject consists of diverse subfields.}
    \label{fig:subjects}
\end{figure*}

In terms of language distribution, 81.8\% of the questions are in English and 18.2\% in Chinese, enabling multilingual research. Question lengths can reach up to 495 words with an average of 77.2 words, reflecting the diversity and richness of the dataset. Explanations average 102.4 words and can extend up to 782 words, providing both concise justifications and detailed elaborations depending on context. In addition, the dataset includes 997 easy, 1,240 medium, and 1,017 hard questions, ensuring a balanced difficulty distribution.

\begin{table*}[t!] %
  \centering
  
  \resizebox{\linewidth}{!}{
  \begin{tabular}{lcccccccc}
    \toprule
  \textbf{Name}   & \textbf{\#Q} & \textbf{\#I} &\textbf{AvgQ} &\textbf{AvgS}& \textbf{\#Subjects} & \textbf{\#Subfields} &\textbf{Level}  & \textbf{Lang} \\
\midrule
ScienceQA\citep{lu2022learn}&21,208&10,332&12.1&47.7&5&41& K-12& en \\
MathVista\citep{lu2024mathvista}&6,141&5,487&15.6&-&1&12& K-12, college& en \\
MATH-V\citep{wang2024measuring}&3,040&3,472&42.3&-&1&16& K-12, college& en \\
MMMU\citep{Yue_2024_CVPR}&11,550&11,264&59.3&108.0&5&39&college& en \\
EMMA\citep{hao2025can}&2,788&2,599&-&-&3&14&K-12, college& en\\
\midrule
\textbf{SciVQR (Ours)}&3,254&3,254&77.2&102.4&6&54&high school, college, graduate& en, zh \\
  \bottomrule
\end{tabular}}
\caption{Statistics for SciVQR and comparisons with existing datasets. \#Q: number of questions, \#I: number of images, AvgQ: average question length;  AvgS: average solution length. \#Subjects: number of scientific subjects. \#Subfields: number of subfields across different scientific subjects.}
\label{tab:comparisons}

  \vspace{15pt} %

  \small
  \resizebox{\linewidth}{!}{
  \begin{tabular}{lccccccc}
    \toprule
    \textbf{Models} & \textbf{Math} & \textbf{Physics}& \textbf{Chemistry} & \textbf{Biology} & \textbf{Geography} & \textbf{Astronomy} & \textbf{Overall}\\
    \midrule
    
    \multicolumn{8}{l}{\textbf{\emph{Open-source Models}}}\\
    LLaVA-NeXT-7B \citep{liu2024llavanext} & 8.0 & 12.6 & 14.1 & 28.9 & 24.2 & 29.0 & 19.5 \\
    LLaVA-OneVision-7B \citep{li2024llava} & 21.9 & 21.0 & 24.6 & 40.6 & 49.2 & 37.5 & 32.5 \\
    MiniCPM-o-2.6-8B \citep{yao2024minicpm}& 26.8 & 23.0 & 22.7 & 35.3 & 38.1 & 27.0 & 28.8 \\
    Molmo-7B-D-0924 \citep{deitke2024molmo}& 20.3 & 19.9 & 23.4 & 38.1 & 41.8 & 35.0 & 29.8 \\
    InternVL3-8B-Instruct \citep{zhu2025internvl3}& 36.4 & 30.9 & 38.4 & 50.3 & 52.9 & 52.5 & 43.6 \\
    InternVL3-14B-Instruct \citep{zhu2025internvl3}& 36.9 & 39.2 & 49.6 & 57.2 & 60.7 & 55.5 & 49.9 \\
    InternVL3-78B-Instruct \citep{zhu2025internvl3}& 47.7 & 46.7 & 52.0 & 60.6 & 57.0 & 57.5 & 53.6 \\
   Qwen2.5-VL-7B-Instruct \citep{bai2025qwen2}& 35.4 & 32.4 & 43.2 & 55.8 & 55.1 & 48.0 & 45.0 \\
   Qwen2.5-VL-72B-Instruct \citep{bai2025qwen2}& 54.3 & 51.8 & 66.6 & 65.3 & 70.3 & 62.5 & 61.8 \\
   Keye-VL-1.5-8B \citep{yang2025kwaikeyevl15technical}& 61.1 & 55.2 & 53.9 & 62.2 & 64.4 & 61.5 & 59.7 \\
   MiMo-VL-7B-RL-2508 \citep{xiaomi2025mimo}& \textbf{65.5} & \textbf{67.5} & \textbf{74.7} & \textbf{71.1} & \textbf{70.6} & \textbf{66.0} & \textbf{69.2} \\
   \midrule
   \multicolumn{8}{l}{\textbf{\emph{Proprietary Models}}}\\
    GPT-4o \citep{hurst2024gpt}& 47.4 & 50.3 & 60.1 & 72.2 & 67.8 & 66.5 & 60.7 \\
    o1 \citep{jaech2024openai}& 79.2 & 73.0 & 79.5 & 80.6 & 76.2 & 74.5 & 77.2 \\
    o4-mini \citep{openai2024systemcard}& \textbf{84.5} & 77.7 & 85.4 & \textbf{80.8} & \textbf{76.5} & 71.0 & \textbf{79.3} \\
    Gemini-2.5-Flash \citep{team2023gemini}& 83.2 & \textbf{78.1} & \textbf{87.6} & 78.6 & 76.2 & \textbf{72.0} & \textbf{79.3} \\
  \bottomrule
\end{tabular}}
\caption{Comparison of model performances across various subjects. The best-performing model in each category is highlighted in \textbf{bold}.}
\label{tab:overall_results}
\end{table*} %

\paragraph{Domain Diversity.} Each question belongs to one of six subjects: mathematics, physics, chemistry, astronomy, geography, and biology. Within each subject, questions are categorized into fine-grained types (e.g., Arithmetic, Logic, Optics, Cosmology). SciVQR covers 6 subjects and 54 categories. The treemap in Figure~\ref{fig:subjects} visualizes this distribution, showing that questions span a wide and diverse range of scientific domains.

\subsection{Comparisons with Existing Benchmarks}
To advance beyond existing limitations, SciVQR introduces critical innovations across four dimensions, as shown in Table~\ref{tab:comparisons}. 
In disciplinary coverage, we encompass 6 subjects and 54 subfields across scientific domains, offering broader coverage compared to existing benchmarks such as ScienceQA (5 subjects, 41 subfields) and MMMU (5 subjects, 39 subfields).
In difficulty design, we establish a multi-tiered structure ranging from high school to graduate-level multi-hop reasoning, avoiding the saturation of datasets restricted to K-12 or college levels.
In reasoning transparency, we pioneer stepwise solution annotations that trace intermediate reasoning steps. This enables fine-grained evaluation of CoT reasoning across multiple dimensions (faithfulness, informativeness, redundancy, hallucination, and missing steps), moving far beyond prevalent answer-only assessments.
In linguistic inclusivity, we support dual-language (English/Chinese) evaluation to mitigate monolingual bias and enable cross-lingual reasoning studies, whereas benchmarks like ScienceQA and MMMU are limited to English only.
Together, these features promote genuine multimodal reasoning and set a new standard for comprehensive scientific evaluation.

\section{Experiments}
\subsection{Experimental Setup}

\textbf{Baselines.} We selected representative multimodal large language models as baselines for the SciVQR benchmark, listed below.

\textit{Open-source models:} These include Qwen2.5-VL-72B \citep{bai2025qwen2}, LLaVA-OneVision-7B \citep{li2024llava}, LLaVA-NeXT-7B \citep{liu2024llavanext}, InternVL3-8B \citep{zhu2025internvl3}, and MiMo-VL-7B-RL-2508 \citep{xiaomi2025mimo}, among others. These models represent the state-of-the-art in publicly available multimodal LLMs across various parameter scales and training strategies, offering a broad view of open-source capabilities in scientific visual question answering.

\textit{Proprietary models:} These include GPT-4o \citep{hurst2024gpt}, o1 \citep{jaech2024openai}, o4-mini \citep{openai2024systemcard}, and Gemini-2.5-Flash (gemini-2.5-flash-preview-04-17) \citep{team2023gemini}. Compared to open-source models, these proprietary models typically show stronger cross-modal understanding and reasoning capabilities, reflecting the latest advances in multimodal reasoning. Models such as o1 and o4-mini are optimized for complex reasoning tasks with targeted training strategies, while Gemini-2.5-Flash supports both reasoning-enabled and reasoning-disabled modes. We used the reasoning-enabled mode in our evaluation.

All models were evaluated under a zero-shot setting to assess their ability to answer questions without fine-tuning or few-shot examples. 

For prompting, we designed two formats: one presenting direct questions, and another incorporating CoT reasoning to study its effect on understanding and inference. Results in Table~\ref{tab:overall_results} are reported under the CoT setting. In Section~\ref{sec:cot_prompt}, we further investigate the impact of CoT prompting by comparing model performance with and without CoT prompts. The corresponding prompt templates are provided in Appendix~\ref{sec:prompts}.

All experiments were conducted on an internal computing cluster equipped with NVIDIA A800 GPUs.

\begin{table*}
  \centering
  \resizebox{\linewidth}{!}{
  \begin{tabular}{l|c|cccccc|c}
    \toprule
    \textbf{Models} & \textbf{CoT} &\textbf{Math} & \textbf{Physics}& \textbf{Chemistry} & \textbf{Astronomy} & \textbf{Geography} & \textbf{Biology} & \textbf{Overall}\\
    \midrule
     \multirow{3}*{InternVL3-8B-Instruct  \citep{zhu2025internvl3}} & $\times$ & 34.1 & 32.7 & 39.9 & 51.4 & 58.8 & 51.0 & 44.7 \\
     & $\checkmark$ & 36.4 & 30.9 & 38.4 & 50.3 & 52.9 & 52.5 & 43.6 \\
     &  & 
\textcolor{red}{+2.3} & \textcolor{OliveGreen}{-1.8} & \textcolor{OliveGreen}{-1.5} & \textcolor{OliveGreen}{-1.1} & \textcolor{OliveGreen}{-5.9} & \textcolor{red}{+1.5} & \textcolor{OliveGreen}{-1.1} \\
    \midrule
     \multirow{3}*{Qwen2.5-VL-72B-Instruct \citep{bai2025qwen2}} & $\times$ & 39.2 & 42.4 & 55.4 & 62.2 & 67.5 & 56.5 & 53.9 \\
     & $\checkmark$ & 54.3 & 51.8 & 66.6 & 65.3 & 70.3 & 62.5 & 61.8 \\
     &  & \textcolor{red}{+15.1} & \textcolor{red}{+9.4} & \textcolor{red}{+11.2} & \textcolor{red}{+3.1} & \textcolor{red}{+2.8} & \textcolor{red}{+6.0} & \textcolor{red}{+7.9} \\
    \midrule
    \multirow{3}*{GPT-4o \citep{hurst2024gpt}} & $\times$ & 33.9 & 42.6 & 52.5 & 68.9 & 68.1 & 61.5 & 54.6 \\
     & $\checkmark$ & 47.4 & 50.3 & 60.1 & 72.2 & 67.8 & 66.5 & 60.7 \\
     &  & \textcolor{red}{+13.5} & \textcolor{red}{+7.7} & \textcolor{red}{+7.6} & \textcolor{red}{+3.3} & \textcolor{OliveGreen}{-0.3} & \textcolor{red}{+5.0} & \textcolor{red}{+6.1} \\
    \midrule
     \multirow{3}*{Keye-VL-1.5-8B \citep{yang2025kwaikeyevl15technical}}& $\times$ & 29.9 & 31.3 & 42.0 & 52.5 & 59.1 & 45.5 & 43.4 \\
     & $\checkmark$ & 61.1 & 55.2 & 53.9 & 62.2 & 64.4 & 61.5 & 59.7 \\
     &  & \textcolor{red}{+31.2} & \textcolor{red}{+24.1} & \textcolor{red}{+11.9} & \textcolor{red}{+9.7} & \textcolor{red}{+5.3} & \textcolor{red}{+16.0} & \textcolor{red}{+16.3} \\

  \bottomrule
\end{tabular}}
\caption{Influence of CoT on model performances across various subjects. For Keye-VL-1.5-8B, ``CoT on/off'' corresponds to enabling or disabling its internal reasoning mode.}
\label{tab:cot}
\end{table*}

\subsection{Overall Results}

Table~\ref{tab:overall_results} presents the performance of various multimodal large language models on the SciVQR benchmark across different scientific subjects.

\textbf{Comparison between proprietary and open-source models:} Proprietary models markedly surpass open-source models in overall accuracy. GPT-4o, o1, o4-mini, and Gemini-2.5-Flash all achieve accuracy rates above 60\%, with o4-mini and Gemini-2.5-Flash achieving the highest overall score of 79.3\%. In contrast, among open-source models, Qwen2.5-VL-72B ranks highest (61.8\%). Although reasoning-oriented models such as Keye-VL-1.5-8B (59.7\%) and MiMo-VL-7B-RL-2508 (69.2\%) show clear improvements over previous open-source baselines, their performance still lags notably behind proprietary reasoning models, suggesting that closed-source reasoning optimization remains more advanced. Lightweight models such as LLaVA-NeXT-7B and LLaVA-OneVision-7B remain below 40\%, indicating their limitations in complex scientific reasoning.

\textbf{Importance of reasoning capability:} Models with strong reasoning ability such as o1 and o4-mini achieve the best results across all subjects (77.2\% and 79.3\%), demonstrating that scientific problem-solving requires not only perception but also multi-step reasoning. The strong performance of MiMo-VL-7B-RL-2508 among open-source models further confirms that reasoning optimization, rather than scale, is key to fidelity and completeness. The superior performance of reasoning-oriented models highlights the importance of explicit reasoning mechanisms in multimodal model design for handling complex and abstract scientific tasks.

\textbf{Performance trends across subjects:} Model performance varies notably across subjects. Mathematics and physics remain most challenging, especially for open-source models, due to intensive reasoning and computation requirements. By contrast, biology and geography rely more on conceptual understanding. Models enhanced for reasoning exhibit more consistent performance across subjects. For example, o4-mini achieves over 70\% accuracy in all subject areas.

\begin{table}[t]
\small
\centering
\renewcommand{\arraystretch}{0.97} %
\small
\vspace{10pt}
\begin{tabular}{lccc}
\toprule
\textbf{Models} & \textbf{Easy} & \textbf{Medium} & \textbf{Hard} \\
\midrule
\multicolumn{4}{l}{\textbf{\emph{Open-source Models}}}\\
LLaVA-OneVision-7B & 40.1 & 25.2 & 18.8 \\
InternVL3-8B-Instruct & 59.3 & 37.6 & 22.6 \\
InternVL3-14B-Instruct & 65.1 & 43.0 & 27.2 \\
InternVL3-78B-Instruct & 70.0 & 50.0 & 33.5 \\
Qwen2.5-VL-7B-Instruct & 51.6 & 39.8 & 31.2 \\
Qwen2.5-VL-72B-Instruct & 75.1 & 58.2 & 42.9 \\
Keye-VL-1.5-8B & 73.4 & 59.1 & 45.6 \\
MiMo-VL-7B-RL-2508 & \textbf{79.7} & \textbf{69.7} & \textbf{55.4} \\
\midrule
\multicolumn{4}{l}{\textbf{\emph{Proprietary Models}}}\\
GPT-4o & 69.4 & 53.5 & 44.8 \\
o1 & 87.0 & 77.4 & 67.8 \\
o4-mini & 89.0 & \textbf{82.0} & \textbf{72.0} \\
Gemini-2.5-Flash & \textbf{89.4} & 81.5 & 71.1 \\
\bottomrule
\end{tabular}
\caption{Results across questions of different difficulty levels.}
\label{tab:difficulty}
\end{table}

\textbf{Performance across questions of different difficulty levels:} As shown in Table~\ref{tab:difficulty}, all models perform best on easy questions and worst on hard ones, confirming the dataset's valid difficulty stratification. Open-source models consistently lag behind proprietary ones across all levels, and the performance gap further increases with question difficulty. This indicates that while reasoning-optimized models such as o4-mini and Gemini-2.5-Flash demonstrate strong robustness, open-source models still face substantial challenges in complex multi-step reasoning tasks. Additional analyses on language balance and resampling stability are provided in Appendices~\ref{sec:lang-analysis} and~\ref{sec:robustness}, respectively.

\begin{table*}[!t]
\small
\setlength{\tabcolsep}{3pt} %
  \centering
  \resizebox{\linewidth}{!}{
  \begin{tabular}{lcccccc}
    \toprule
    \textbf{Models} & \textbf{Faithful} & \textbf{Informative} & \textbf{Redundancy} & \textbf{Hallucination} & \textbf{Missing} & \textbf{Overall}\\
    \midrule
     \multicolumn{7}{l}{\textbf{\emph{Open-source Models}}}\\
    LLaVA-NeXT-7B~\citep{liu2024llavanext} & 2.7 & 3.9 & 5.6 & 3.1 & 2.7 & 3.6 \\
     Qwen2.5-VL-7B-Instruct~\citep{bai2025qwen2}& 3.3 & 4.1 & \textbf{7.2} & 3.9 & 3.4 & 4.4 \\
     Qwen2.5-VL-72B-Instruct~\citep{bai2025qwen2}& 4.4 & 5.2 & 6.8 & 5.1 & 4.6 & 5.2 \\
     Keye-VL-1.5-8B \citep{yang2025kwaikeyevl15technical}& 5.5 & 6.2 & 4.3 & 5.5 & 5.5 & 5.4 \\
     MiMo-VL-7B-RL-2508 \citep{xiaomi2025mimo}& \textbf{6.0} & \textbf{6.4} & 3.6 & \textbf{6.1} & \textbf{6.0} & \textbf{5.6} \\
    \midrule
    \multicolumn{7}{l}{\textbf{\emph{Proprietary Models}}}\\
    GPT-4o \citep{hurst2024gpt}& 5.4 & 6.1 & 7.8 & 6.1 & 5.6 & 6.2 \\
    o1 \citep{jaech2024openai}& 6.5 & 7.0 & \textbf{8.6} & 7.3 & 6.6 & 7.2 \\
    o4-mini~\citep{openai2024systemcard} & \textbf{7.1} & \textbf{7.5} & \textbf{8.6} & \textbf{7.8} & \textbf{7.3} & \textbf{7.7} \\
    Gemini-2.5-Flash~\citep{team2023gemini}& 6.2 & 6.7 & 7.3 & 6.9 & 6.4 & 6.8 \\
  \bottomrule
\end{tabular}}
\caption{Results of reasoning quality evaluation.}
\label{tab:reasoning_results}
\end{table*}

\subsection{Does CoT Help in Answering SciVQR Questions?}
\label{sec:cot_prompt}

Table~\ref{tab:cot} compares model performance under two prompting settings: with and without CoT prompting.

For most models, CoT prompting leads to clear improvements. For example, Qwen2.5-VL-72B gains +7.9\% overall, with substantial boosts in mathematics (+15.1\%) and chemistry (+11.2\%). Similarly, GPT-4o benefits from CoT prompting, showing a +6.1\% overall increase, with notable gains in mathematics (+13.5\%) and physics (+7.7\%). These results indicate that CoT prompting can substantially assist models in solving multi-step reasoning and structured scientific problems. 

For the reasoning-specialized Keye-VL-1.5-8B, disabling its built-in reasoning mode causes a substantial performance drop ($-16.3\%$ overall), especially in mathematics ($-31.2\%$) and physics ($-24.1\%$). This degradation reflects the model's strong reliance on CoT-style reasoning, as it is primarily trained under such paradigms.

However, CoT prompting does not always bring benefits. InternVL3-8B shows a slight drop ($-1.1\%$) after applying CoT, suggesting limitations in following complex instruction formats or maintaining concise, well-structured responses. The generation of verbose or off-format responses may hinder the intended benefits of CoT reasoning. 

In summary, CoT prompting generally enhances reasoning for models with strong instruction adherence, while exposing weaknesses in those lacking such capability. These results underscore the importance of robust instruction-following capabilities as a key factor in determining the effectiveness of reasoning-oriented prompting strategies.

\subsection{Reasoning Quality Evaluation}
\label{sec:cot_ablation}

To systematically evaluate the quality of model-generated reasoning, we adopt a structured rubric assessing five core dimensions of CoT quality: \textit{Faithfulness}, \textit{Informativeness}, \textit{Redundancy}, \textit{Hallucination}, and \textit{Missing Steps}:

\begin{itemize}[left=0pt, labelsep=1em]
    \setlength{\itemsep}{3pt}
    \item \textbf{Faithfulness:} Measures alignment between the model's reasoning steps and the ground-truth solution trace.
    \item \textbf{Informativeness:} Assesses how much relevant and critical information is preserved in the model's reasoning.
    \item \textbf{Redundancy:} Evaluates whether reasoning steps are unnecessarily repeated or rephrased without contributing to the logic.
    \item \textbf{Hallucination:} Detects fabricated or irrelevant reasoning steps unsupported by the problem context or solution.
    \item \textbf{Missing Steps:} Identifies whether essential intermediate steps are omitted, affecting the completeness of reasoning.
\end{itemize}

Each dimension is rated on a 10-point scale, and an overall score is computed by averaging the sub-scores. The rubric aligns with human evaluation standards, focusing on step-level fidelity and informativeness rather than final-answer correctness. We employ GPT-4o as an automated judge to compare model-generated CoTs with expert-authored solutions on a diverse SciVQR subset (see Appendix~\ref{sec:reasoning_prompts} for detailed evaluation prompts).

As shown in Table~\ref{tab:reasoning_results}, reasoning-optimized models such as o4-mini (7.7) and o1 (7.2) substantially outperform instruction-tuned models like Qwen2.5-VL-72B (5.2) and Qwen2.5-VL-7B (4.4) across all dimensions. For example, o4-mini exceeds Qwen2.5-VL-72B by +2.5 overall, demonstrating stronger faithfulness, completeness, and groundedness of reasoning. In contrast, scaling Qwen2.5-VL from 7B to 72B yields only a marginal +0.8 improvement despite a tenfold parameter increase.

Notably, the reasoning-finetuned open-source model MiMo-VL-7B-RL-2508 achieves 5.6 overall, surpassing both Qwen2.5-VL variants despite its smaller scale, which highlights that reasoning optimization, not size, drives better CoT quality. Nevertheless, a clear gap remains between open-source and proprietary systems: although MiMo-VL-7B-RL-2508 attains higher task accuracy than GPT-4o, its reasoning is more verbose and repetitive, lowering conciseness and quality. These findings underline the continued advantage of proprietary models in producing coherent, faithful reasoning traces. 

\section{Conclusion}
The introduction of SciVQR represents a significant advancement in evaluating the scientific reasoning capabilities of MLLMs. By encompassing a broad range of scientific subjects, domain-specific visuals, and expert-authored solution traces, SciVQR not only assesses final answers but also scrutinizes the reasoning processes behind them. This unique design provides deeper insight into how models arrive at their conclusions, offering a clearer understanding of the structure and quality of their reasoning.

We contend that strong performance on SciVQR, particularly in solution traceability, should be a critical criterion for evaluating the scientific reasoning capabilities of MLLMs. Therefore, SciVQR will play a pivotal role in advancing multimodal scientific intelligence, driving models toward more transparent, structured, and sophisticated reasoning.

\section*{Limitations}
While SciVQR provides a comprehensive and challenging framework, it does have limitations, including potential biases in the manual curation process and the inherent challenges of capturing the full complexity of scientific reasoning. Moreover, the reliance on expert-authored solutions might introduce subjective factors that could affect the evaluation.

These limitations highlight areas for improvement in future iterations of SciVQR, where addressing biases and ensuring a more objective and comprehensive evaluation framework will be essential to enhance its utility and accuracy.

\section*{Acknowledgements}
This research is supported by the Strategic Priority Research Program of Chinese Academy of Sciences under Grant XDB1350103, the National Natural Science Foundation of China (62437001, 62436001, 62531026), the Key Research Development Program of Jiangsu Province under Grant BE2023016-3, and the Natural Science Foundation of Jiangsu Province under Grant BK20243051.

\bibliography{custom}

\clearpage

\appendix
\section{Language Distribution Analysis}
\label{sec:lang-analysis}
To assess the potential impact of language distribution on model evaluation, we conducted a language-specific performance analysis. Table~\ref{tab:lang-performance} shows the accuracy of selected models on the English and Chinese subsets of SciVQR.
\begin{table}[h]
\centering
\small
\begin{tabularx}{\linewidth}{lcc}
\toprule
Model & English (\%) & Chinese (\%) \\
\midrule
LLaVA-NeXT-7B & 16.4 & 9.1 \\
MiniCPM-o-2.6-8B & 29.3 & 19.1 \\
InternVL3-14B-Instruct & 45.7 & 40.8 \\
GPT-4o & 57.9 & 45.7 \\
Qwen2.5-VL-72B-Instruct & 59.6 & 54.1 \\
o1 & 79.2 & 69.1 \\
o4-mini & 83.0 & 72.2 \\
Gemini-2.5-Flash & 81.5 & 76.9 \\
\bottomrule
\end{tabularx}
\caption{Accuracy of selected models on English and Chinese subsets of SciVQR.}
\label{tab:lang-performance}
\end{table}

The results reveal that most multilingual models exhibit consistent performance trends across both languages. For most models, accuracy on English questions is slightly higher than on Chinese ones. This pattern may arise from differences in the models' multilingual training coverage or optimization, as many models have stronger English capabilities due to larger and more diverse English pretraining corpora.

\section{Robustness under Subsampling and Bootstrapping}
\label{sec:robustness}

To test evaluation stability, we conducted subsampling and bootstrapping analysis (100 trials each). Table~\ref{tab:robustness} shows that standard deviations do not exceed $1.33\%$, confirming the robustness of the evaluation.

\begin{table}[!htbp] %
\centering
\footnotesize
\setlength{\tabcolsep}{3pt}
\begin{tabular}{lcc}
\toprule
Model & Subsample (\%) & Bootstrap (\%) \\
\midrule
LLaVA-NeXT-7B & 0.78 & 0.63 \\
MiniCPM-o-2.6-8B & 1.10 & 0.78 \\
InternVL3-14B-Instruct & 1.33 & 0.87 \\
Qwen2.5-VL-72B-Instruct & 1.22 & 0.84 \\
GPT-4o & 1.24 & 0.89 \\
o1 & 1.05 & 0.73 \\
o4-mini & 1.03 & 0.69 \\
Gemini-2.5-Flash & 0.95 & 0.68 \\
\bottomrule
\end{tabular}
\caption{Standard deviations of model accuracy under subsampling and bootstrapping. ``Subsample'' refers to the standard deviation of model accuracy under subsampling, and ``Bootstrap'' refers to the standard deviation under bootstrapping.}
\label{tab:robustness}
\end{table}

\section{Prompt Templates}
\label{sec:prompts}
\subsection{Prompt for Response Generation}
The prompt for response generation is as follows: 

\begin{tcolorbox}[colback=gray!5, colframe=black!45,
    fonttitle=\bfseries, title= System Prompt used for Response Generation with CoT]
\small

\begin{lstlisting}[breaklines=true, breakindent=0pt]
Please solve the problem step by step and put your answer in one "boxed{}". If it is a multiple choice question, only one letter is allowed in the "boxed{}" 
{question} 
{options}
\end{lstlisting}
\end{tcolorbox}

\begin{tcolorbox}[colback=gray!5, colframe=black!45,
    fonttitle=\bfseries, title= System Prompt used for Response Generation without CoT]
\small

\begin{lstlisting}[breaklines=true, breakindent=0pt]
Answer the question directly without explanations or reasoning and put your answer in one "boxed{}". If it is a multiple choice question, only one letter is allowed in the "boxed{}"  
{question} 
{options}
\end{lstlisting}
\end{tcolorbox}

\subsection{Prompt for Evaluating Open Questions}

The prompt for open questions evaluation is as follows: 

\begin{tcolorbox}[colback=gray!5, colframe=black!45,
    fonttitle=\bfseries, title= System Prompt used for Open Questions Evaluation]
\small

\begin{lstlisting}[breaklines=true, breakindent=0pt]
You are given a response from a model and the correct answer. Your task is to determine if the model's response is correct. You should only return 'true' if the response matches the answer. If the answer is a floating-point number greater than 1, when it is represented in scientific notation, a difference of up to 0.1 is allowed. Otherwise, return 'false'.
Response: {example['response']}
Correct Answer: {example['answer']}
Is the response correct? (true/false)
\end{lstlisting}
\end{tcolorbox}

\subsection{Prompt for Evaluating Reasoning Quality}
\label{sec:reasoning_prompts}
The prompt for reasoning quality evaluation is as follows: 

\begin{tcolorbox}[colback=gray!5, colframe=black!45,
    fonttitle=\bfseries, title= System Prompt used for Reasoning Quality Evaluation (Part 1)]
\small

\begin{lstlisting}[breaklines=true, breakindent=0pt]
You are a reasoning evaluator tasked with analyzing the alignment, clarity, and completeness of reasoning steps in generated text. Your role is to assess the reasoning path between the ground truth and the LLM-generated response based on the criteria below:

** Faithfulness (1-10) :**
Description: Assesses how closely the LLM's reasoning steps reflect those in the source (ground truth).
Scoring Guide:
    9-10: Reasoning steps almost entirely mirror the original.
    7-8: Most reasoning is aligned, with only minor deviations.
    5-6: Partial alignment, but several key differences or omissions.
    3-4: Few reasoning elements match; many are missing or incorrect.
    1-2: Majority of reasoning does not correspond to the source.

** Repetition and Redundancy (1-10) :**
Description: Evaluates repeated or unnecessary reasoning, including paraphrased points that do not add substance.
Scoring Guide:
    9-10: No or minimal redundant content.
    7-8: Slight repetition that does not hinder comprehension.
    5-6: Noticeable redundancy without meaningful contribution.
    3-4: Frequent repetition that disrupts the logical flow.
    1-2: Excessive and disruptive duplication of reasoning steps.

** Informativeness (1-10) :**
Description: Measures whether the reasoning includes all essential details derived from the source.
Scoring Guide:
    9-10: Nearly all important information is captured and accurately conveyed.
    7-8: Most major points are included; only minor items are overlooked.
    5-6: Some significant pieces of information are lacking or underdeveloped.
    

\end{lstlisting}
\end{tcolorbox}

\begin{tcolorbox}[colback=gray!5, colframe=black!45,
    fonttitle=\bfseries, title= System Prompt used for Reasoning Quality Evaluation (Part 2)]
\small
\begin{lstlisting}[breaklines=true, breakindent=0pt]
    3-4: Many critical points are missing.
    1-2: Very poor coverage of key information.
    
** Hallucination (1-10) :**
Description: Identifies reasoning steps that are fabricated or unsupported by the source material.
Scoring Guide:
    9-10: All reasoning is well-grounded in the original content.
    7-8: One or two minor hallucinations are present.
    5-6: Several steps introduce irrelevant or false information.
    3-4: Many hallucinated steps, with partial grounding.
    1-2: Reasoning is largely based on invented content.

** Missing Step (1-10) :**
Description: Checks whether any necessary components of the reasoning chain are absent.
Scoring Guide:
    9-10: No critical reasoning steps are missing.
    7-8: Only minor omissions that don't impact the overall conclusion.
    5-6: Several important steps are left out, affecting the output's quality.
    3-4: Multiple essential steps are not included.
    1-2: Major reasoning gaps make the logic incomplete.

Consistency and Evaluation Guidelines:
    1. Adhere strictly to the above scoring scales.
    2. Always re-read both the ground truth and the LLM response thoroughly before scoring.
    3. Perform a direct comparison of the reasoning sequences to assess agreement or deviation.
    4. Use reference examples, if provided, to maintain evaluation consistency.
    5. Avoid subjective judgment; stay within the defined thresholds.
    6. After assigning scores for each category, compute the final Overall score as the average.
\end{lstlisting}
\end{tcolorbox}

\begin{tcolorbox}[colback=gray!5, colframe=black!45,
    fonttitle=\bfseries, title= System Prompt used for Reasoning Quality Evaluation (Part 3)]
\small

\begin{lstlisting}[breaklines=true, breakindent=0pt]   

Output must be returned in the following Python dictionary format exactly as shown — do not add or modify anything, as this output feeds directly into a processing pipeline and any deviation will cause a system failure:

# Example output: {'Faithfulness': 8.0, 'Informativeness': 8.5, 'Repetition&Redundancy': 9.0, 'Hallucination': 9.5, 'Missing': 8.5, 'Overall': 8.65}

\end{lstlisting}
\end{tcolorbox}

\vspace{10pt}
\section{Performance on Multiple-Choice and Direct-Response Questions}
\label{sec:mc_dr_accuracy}

To provide a more granular understanding of model capabilities, we report separate performance on multiple-choice (MC) and direct-response (DR) questions. Results for a representative subset of models are shown in Table~\ref{tab:mc_dr_performance}.

\paragraph{Overall Trends:}
\begin{itemize}[leftmargin=*]
    \item \textbf{Higher accuracy for MC than DR.} All models perform better on MC questions, confirming that open-ended reasoning and generation are inherently more challenging than option selection.
    \item \textbf{Larger gaps for mid-scale open-source models.} Models such as Qwen2.5-VL-7B and InternVL3-8B show substantial performance drops from MC to DR, indicating weaker robustness in free-form reasoning.
    \item \textbf{Smaller gaps for stronger models.} Keye-VL-1.5-8B and GPT-4o exhibit more stable DR performance, suggesting better generalization beyond fixed-choice settings.
\end{itemize}

\paragraph{Model-wise Observations:}
\begin{itemize}[leftmargin=*]
    \item Qwen2.5-VL-72B-Instruct demonstrates notable strengths in chemistry and geography, achieving the highest MC and DR scores among the listed models in these subjects (e.g., 67.7 MC / 61.3 DR in chemistry; 70.0 MC / 71.7 DR in geography).
    \item Keye-VL-1.5-8B performs particularly well in mathematics and physics, showing competitive or leading results in these areas (e.g., 61.2 MC / 60.3 DR in math; 50.1 MC / 65.5 DR in physics).
\end{itemize}

\begin{table*}[t] %
\centering
\small %
\setlength{\tabcolsep}{4.5pt} %
\begin{tabular}{@{} lcccccccccccccc @{}}
\toprule
& \multicolumn{2}{c}{\textbf{Math}} & \multicolumn{2}{c}{\textbf{Physics}} & \multicolumn{2}{c}{\textbf{Chem.}} & \multicolumn{2}{c}{\textbf{Biology}} & \multicolumn{2}{c}{\textbf{Geo.}} & \multicolumn{2}{c}{\textbf{Astro.}} & \multicolumn{2}{c}{\textbf{Overall}} \\
\cmidrule(lr){2-3} \cmidrule(lr){4-5} \cmidrule(lr){6-7} \cmidrule(lr){8-9} \cmidrule(lr){10-11} \cmidrule(lr){12-13} \cmidrule(lr){14-15}
\textbf{Model} & \textbf{MC} & \textbf{DR} & \textbf{MC} & \textbf{DR} & \textbf{MC} & \textbf{DR} & \textbf{MC} & \textbf{DR} & \textbf{MC} & \textbf{DR} & \textbf{MC} & \textbf{DR} & \textbf{MC} & \textbf{DR} \\
\midrule
InternVL3-8B-Instruct  & 37.1 & 32.9 & 34.1 & 24.6 & 39.5 & 33.3 & 52.5 & 38.6 & 54.2 & 45.7 & 65.0 & 23.3 & 47.1 & 33.1 \\
InternVL3-78B-Instruct & 48.5 & 44.3 & 48.9 & 42.5 & 54.1 & 42.7 & 62.7 & 49.1 & 57.0 & 56.5 & 71.4 & 25.0 & 57.1 & 39.6 \\
Qwen2.5-VL-7B-Instruct & 35.9 & 32.9 & 35.7 & 25.8 & 44.5 & 37.3 & 57.8 & 45.6 & 54.5 & 58.7 & 60.7 & 18.3 & 48.2 & 36.4 \\
Qwen2.5-VL-72B-Instruct& 56.5 & 44.3 & \textbf{55.7} & 44.0 & \textbf{67.7} & \textbf{61.3} & 67.7 & 52.6 & \textbf{70.0} & \textbf{71.7} & 77.1 & 28.3 & \textbf{65.8} & 50.4 \\
Keye-VL-1.5-8B         & \textbf{61.2} & \textbf{60.3} & 50.1 & \textbf{65.5} & 54.9 & 49.3 & 64.0 & 52.6 & 64.3 & 65.2 & 66.4 & \textbf{50.0} & 60.2 & \textbf{54.6} \\
GPT-4o                 & 48.0 & 44.7 & 50.8 & 49.2 & 62.5 & 49.3 & \textbf{73.3} & \textbf{66.7} & 68.6 & 63.0 & \textbf{81.4} & 31.7 & 64.1 & 50.8 \\
\bottomrule
\end{tabular}
\caption{Comparison of Multiple-Choice (MC) and Direct-Response (DR) accuracy across scientific subjects.}
\label{tab:mc_dr_performance}
\end{table*}

\vspace{10pt}
\section{Reasoning Quality across Difficulty Levels}
\label{sec:reasoning_diff}

Reasoning evaluation is a central focus of our work. While Section~\ref{sec:cot_ablation} reports the overall quantitative evaluation of model-generated CoTs using our five-dimension rubric (faithfulness, informativeness, redundancy, hallucination, and missing steps), we additionally include an expanded evaluation here to examine how reasoning performance varies with question difficulty. Table~\ref{tab:reasoning_diff_table} reports the results across Easy, Medium, and Hard questions. 

The results demonstrate the following trends:

\begin{itemize}[leftmargin=*]
    \item Reasoning-optimized models achieve higher scores across all dimensions, showing better alignment with expert solution traces.
    \item Scores decrease progressively from lower- to higher-difficulty questions, indicating that complex problems lead to less faithful and complete CoTs.
    \item This trend supports the reliability of our difficulty annotations and demonstrates that SciVQR captures meaningful differences in reasoning complexity.
\end{itemize}

\begin{table}[h]
\centering
\footnotesize
\setlength{\tabcolsep}{1pt} %
\begin{tabular*}{\linewidth}{@{\extracolsep{\fill}} lccccccc @{}} %
\toprule
\textbf{Model} & \textbf{Diff.} & \textbf{Fai.} & \textbf{Inf.} & \textbf{Red.} & \textbf{Hal.} & \textbf{Mis.} & \textbf{Over.} \\
\midrule
Qwen2.5-VL & Easy & 4.5 & 5.3 & 7.4 & 5.2 & 4.7 & 5.4 \\
(7B-Inst.) & Med. & 4.3 & 5.2 & 7.3 & 5.0 & 4.5 & 5.3 \\
           & Hard & 4.0 & 4.9 & 7.1 & 4.6 & 4.1 & 4.9 \\
\midrule
Keye-VL    & Easy & 6.2 & 6.8 & 5.1 & 6.3 & 6.2 & 6.1 \\
(1.5-8B)   & Med. & 5.5 & 6.2 & 4.2 & 5.4 & 5.4 & 5.4 \\
           & Hard & 5.1 & 5.8 & 3.9 & 5.0 & 5.0 & 5.0 \\
\midrule
GPT-4o     & Easy & 5.9 & 6.5 & 8.0 & 6.7 & 6.0 & 6.6 \\
           & Med. & 5.4 & 6.1 & 7.8 & 6.2 & 5.6 & 6.2 \\
           & Hard & 5.0 & 5.8 & 7.7 & 5.7 & 5.2 & 5.9 \\
\midrule
o4-mini    & Easy & \textbf{7.0} & \textbf{7.4} & \textbf{8.5} & \textbf{7.8} & \textbf{7.1} & \textbf{7.6} \\
           & Med. & 6.4 & 7.0 & 8.4 & 7.2 & 6.6 & 7.1 \\
           & Hard & 6.0 & 6.6 & 8.2 & 6.8 & 6.2 & 6.7 \\
\bottomrule
\end{tabular*}
\caption{Results of reasoning quality evaluation by difficulty level. Abbreviations: Diff. (Difficulty), Fai. (Faithfulness), Inf. (Informativeness), Red. (Redundancy), Hal. (Hallucination), Mis. (Missing), Over. (Overall). Qwen2.5-VL and Keye-VL refer to the Qwen2.5-VL-7B-Instruct and Keye-VL-1.5-8B versions, respectively.}
\label{tab:reasoning_diff_table}
\end{table}

\vspace{10pt}
\section{Error Analysis}
\label{sec:error-analysis}

To analyze the limitations of current large-scale multimodal models, we perform both quantitative and qualitative error analyses based on empirical results. In our study, error instances are classified into categories including Image Perception Error, Refusal to Answer, and Lack of Knowledge, alongside an overarching reasoning category. To further disentangle reasoning bottlenecks, we subdivide reasoning errors into four fine-grained categories: Reasoning Inference Error, Reasoning Premise Error, Reasoning Misused Evidence, and Reasoning Step-wise Omission.

\subsection{Quantitative Error Distribution}
We conducted a quantitative error-type analysis to better understand model failure patterns. Table~\ref{tab:error_type_dist} reports the distribution of error types from 100 randomly sampled incorrect predictions for four representative models. 

\begin{table}[h]
\centering
\footnotesize %
\setlength{\tabcolsep}{1pt} %
\begin{tabular*}{\linewidth}{@{\extracolsep{\fill}} lccccccc @{}} %
\toprule
& \multicolumn{4}{c}{\textbf{Reasoning}} & & & \\
\cmidrule(lr){2-5}
\textbf{Model} & \textbf{Inf.} & \textbf{Pre.} & \textbf{Evi.} & \textbf{Omi.} & \textbf{Per.} & \textbf{Ref.} & \textbf{Kno.} \\
\midrule
Qwen2.5 & 14 & 43 & 6 & 8 & 28 & 1 & 0 \\
GPT-4o & 14 & 37 & 14 & 10 & 23 & 0 & 2 \\
MiMo & 49 & 15 & 4 & 8 & 23 & 0 & 1 \\
o4-mini & 29 & 15 & 15 & 24 & 16 & 0 & 1 \\
\bottomrule
\end{tabular*}
\caption{Error-type distribution (counts) from 100 sampled incorrect cases. Categories include Inf. (Reasoning Inference), Pre. (Reasoning Premise), Evi. (Misused Evidence), Omi. (Step-wise Omission), Per. (Image Perception), Ref. (Refusal), and Kno. (Knowledge). Qwen2.5 and MiMo denote the Qwen2.5-VL-7B-Instruct and MiMo-VL-7B-RL-2508 models, respectively.}
\label{tab:error_type_dist}
\end{table}

Based on the quantitative results, we identify three key observations:

\begin{itemize}[leftmargin=*]
    \item \textbf{Reasoning errors dominate overall.} Across all evaluated models, reasoning failures remain the primary source of errors. This indicates that models frequently struggle to construct complete and coherent reasoning chains even when visual perception is relatively accurate.
    \item \textbf{Standard models show higher premise-level errors.} Models without explicit Long CoT capabilities (e.g., Qwen2.5-VL and GPT-4o) exhibit significantly more Reasoning Premise Errors. Without a structured reasoning process to re-evaluate initial assumptions, their early grounding mistakes directly lead to wrong answers.
    \item \textbf{Reasoning models show higher inference-level errors.} Models that produce explicit reasoning traces (e.g., MiMo-VL and o4-mini) typically ground visual premises correctly. However, their extended reasoning chains introduce more opportunities for logical inconsistencies and incorrect step transitions, making Reasoning Inference Errors their dominant failure mode.
\end{itemize}

\subsection{Qualitative Error Examples}
\begin{figure*}[h]
    \centering
    \includegraphics[width=0.88\linewidth]{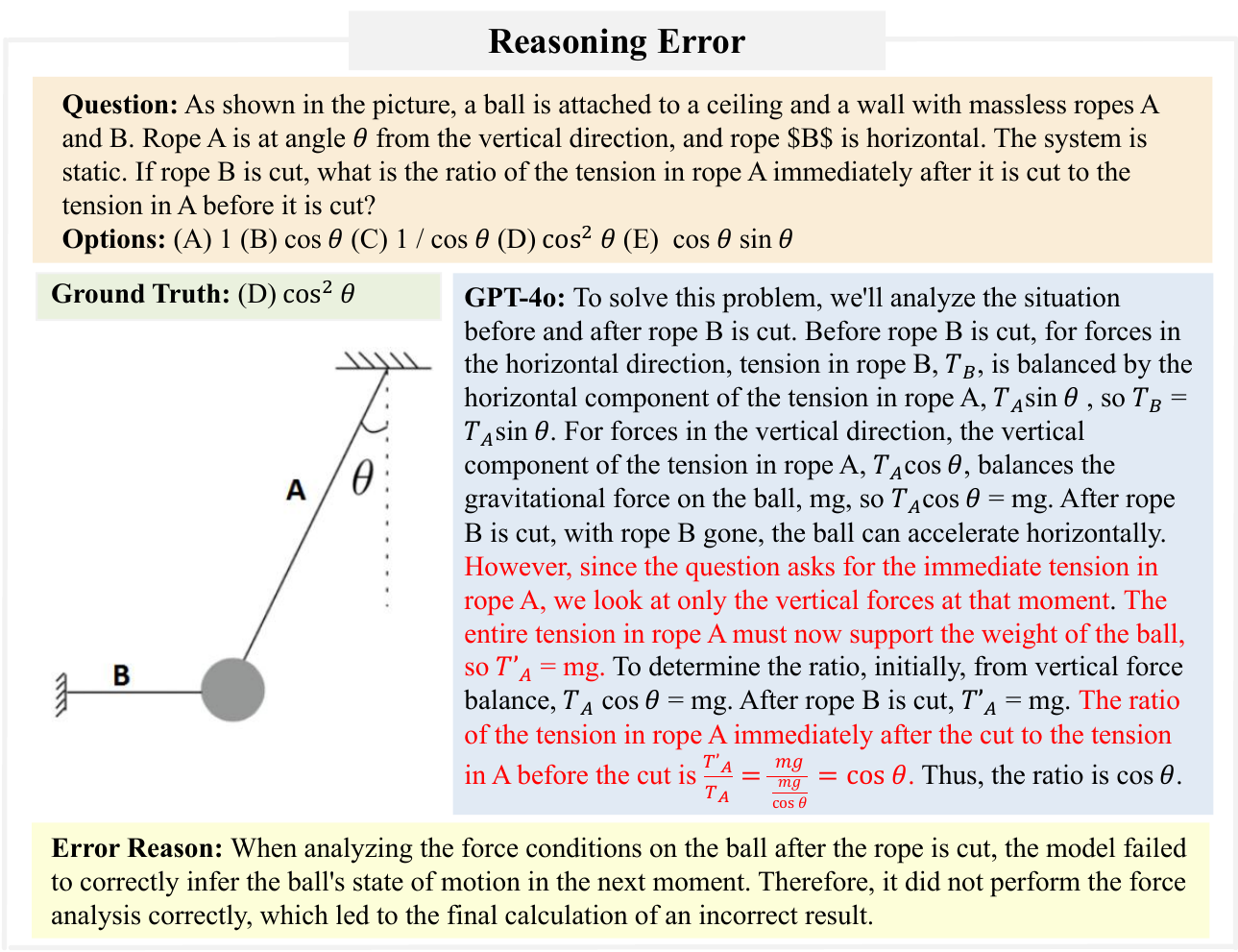}
    \caption{Reasoning error case of GPT-4o.}
    \label{fig:error1}
\end{figure*}

\begin{figure*}[h]
    \centering
    \includegraphics[width=0.88\linewidth]{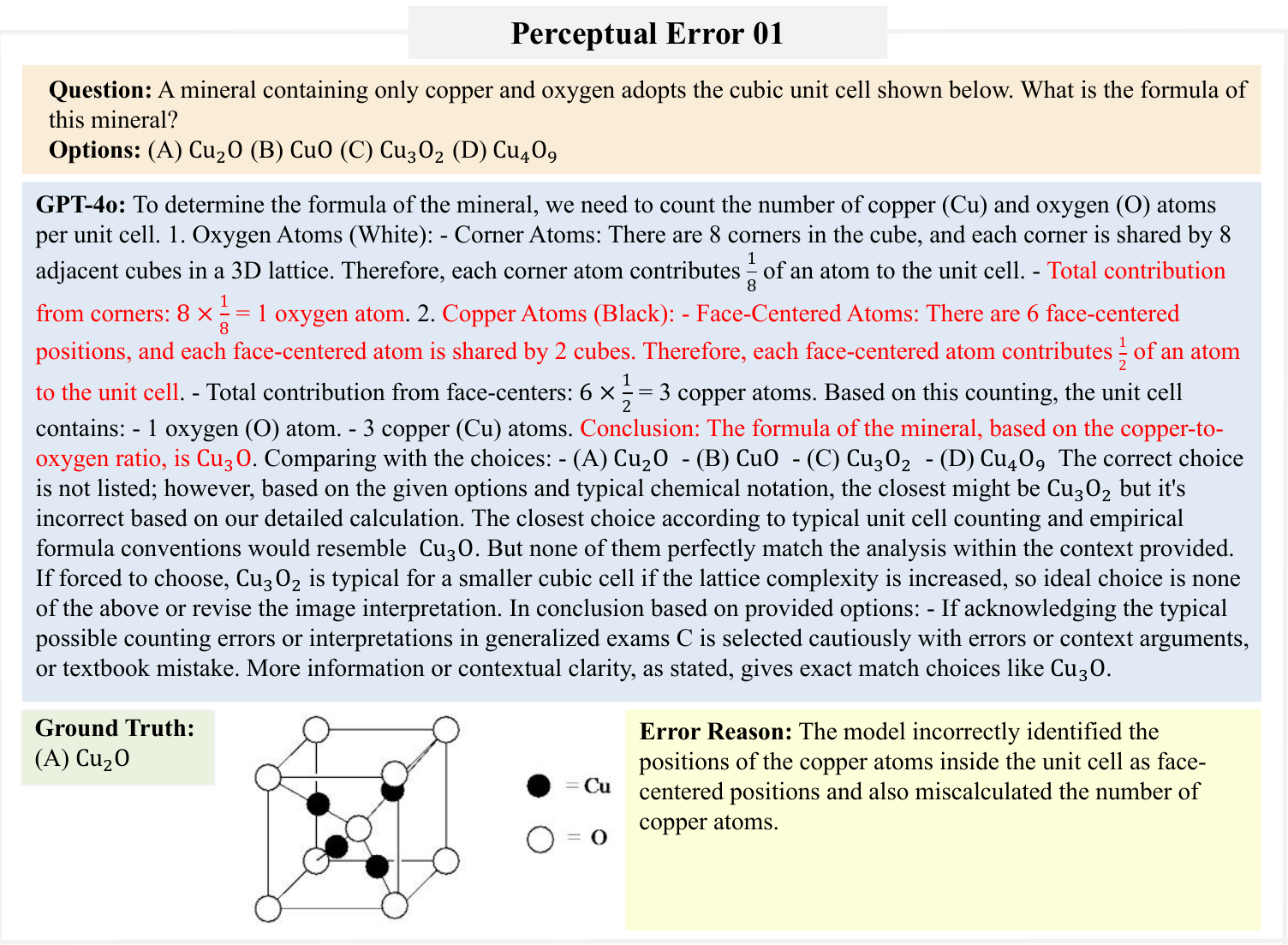}
    \caption{Perception error case of GPT-4o.}
    \label{fig:error2}
\end{figure*}

\begin{figure*}[h]
    \centering
    \includegraphics[width=0.88\linewidth]{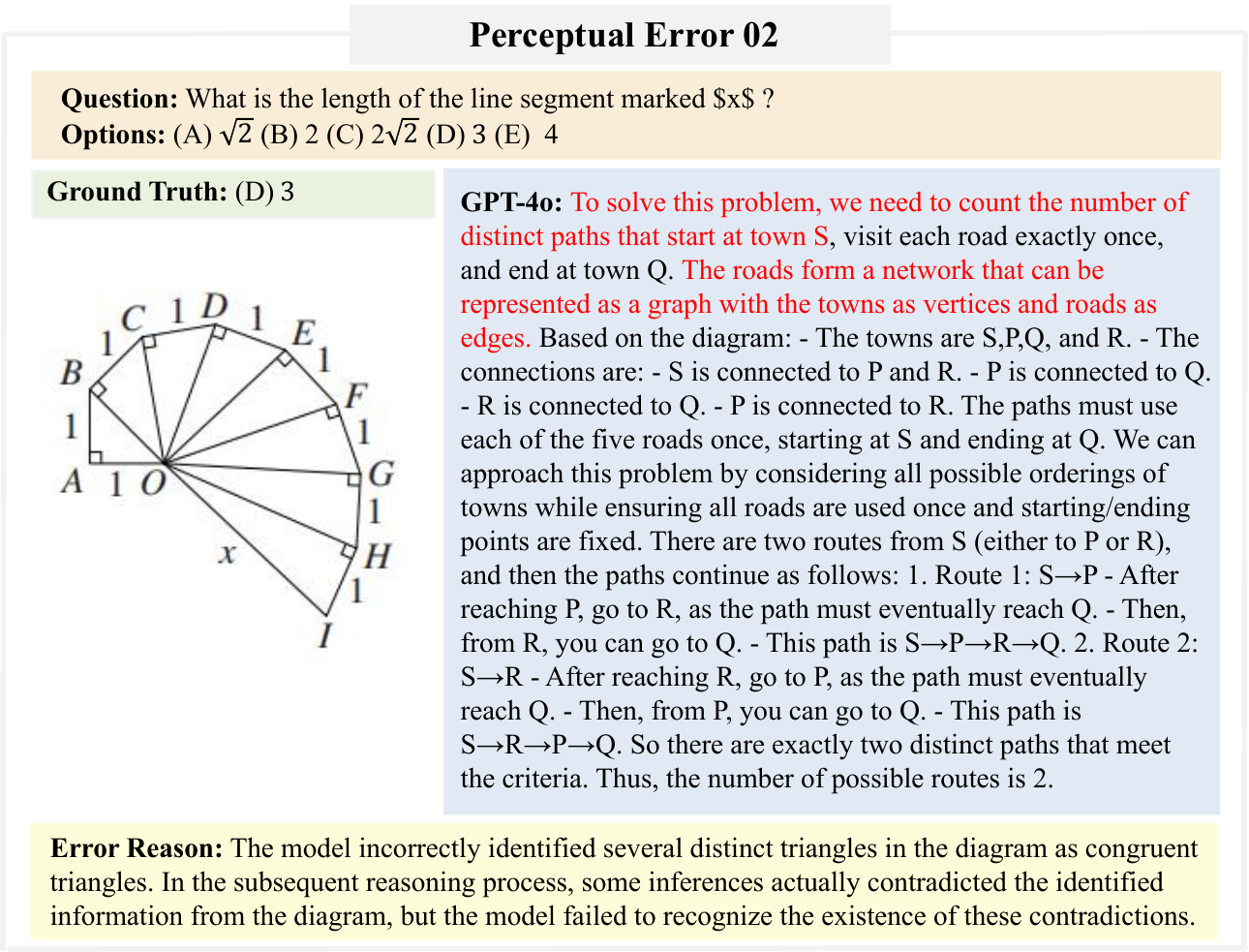}
    \caption{Perception error case of GPT-4o.}
    \label{fig:error3}
\end{figure*}

\begin{figure*}[h]
    \centering
    \includegraphics[width=0.88\linewidth]{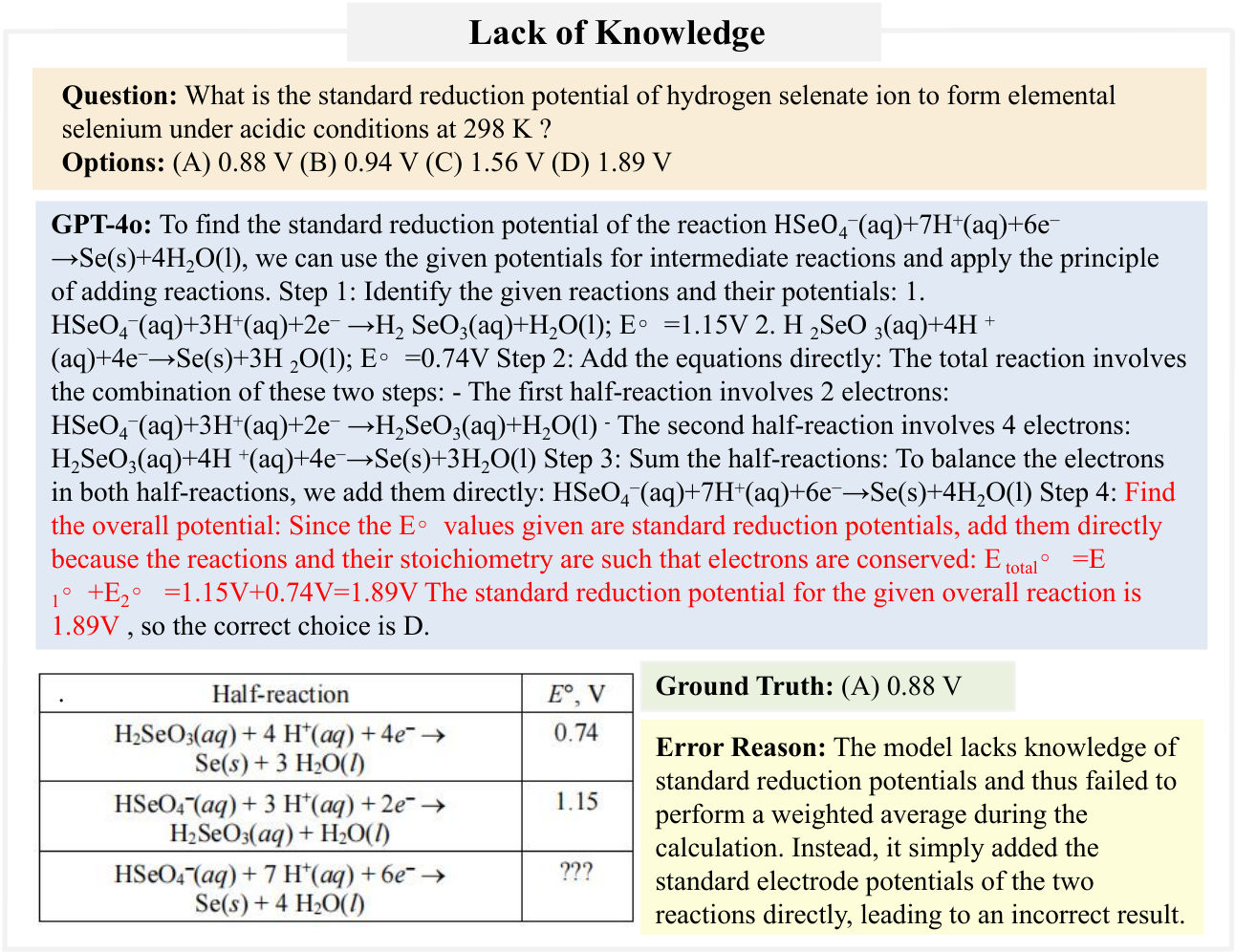}
    \caption{Knowledge gap error case of GPT-4o.}
    \label{fig:error4}
\end{figure*}

\begin{figure*}[h]
    \centering
    \includegraphics[width=0.88\linewidth]{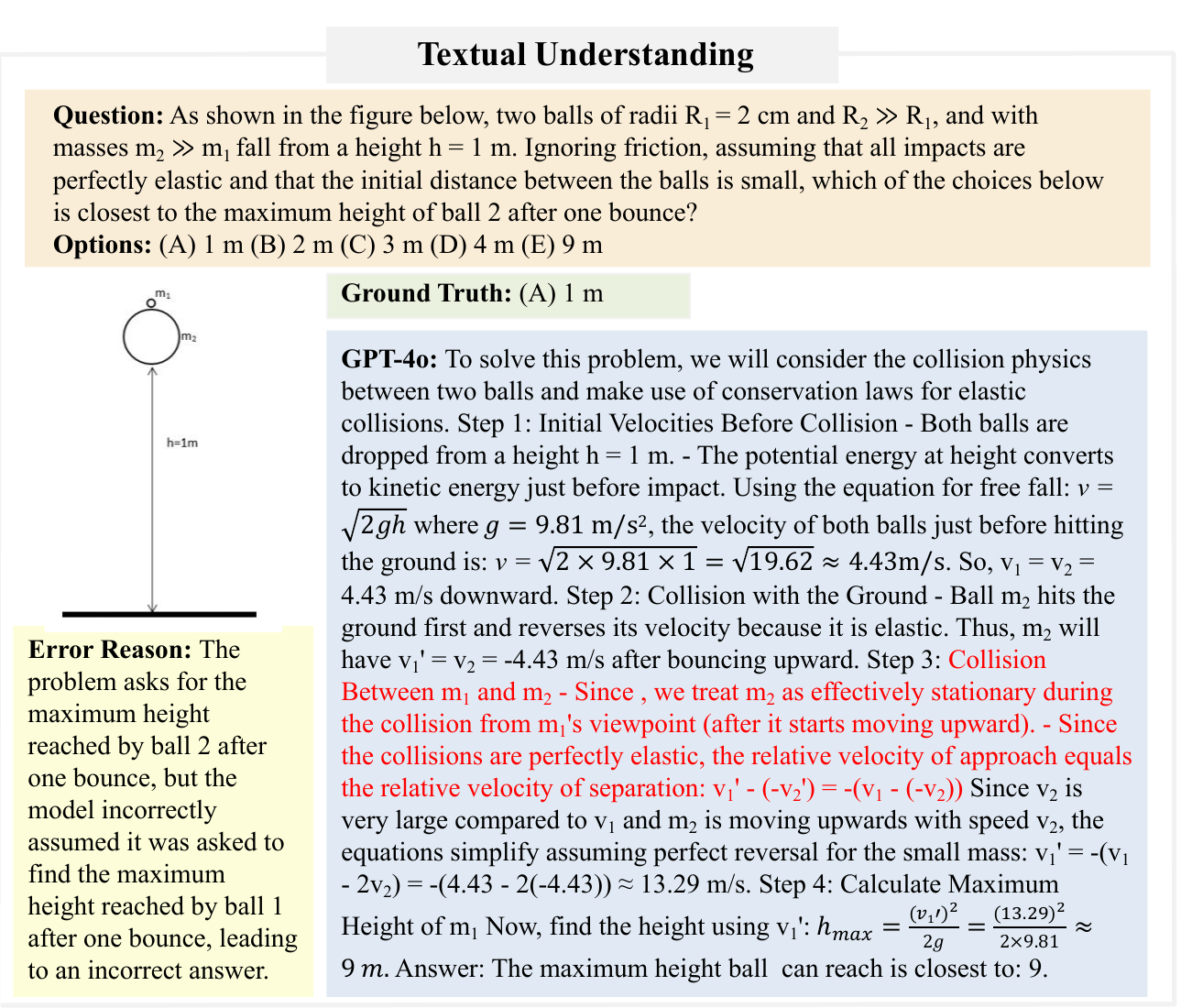}
    \caption{Textual understanding error case of GPT-4o.}
    \label{fig:error5}
\end{figure*}

Following the quantitative breakdown, we provide concrete qualitative examples to illustrate these specific limitations. Figures \ref{fig:error1} to \ref{fig:error5} present typical failure cases of GPT-4o, spanning Reasoning Errors, Image Perception Errors, and Lack of Knowledge.

\clearpage
\onecolumn
\section{More Examples of the Dataset}

\vspace{120pt}
\begin{figure}[h]
    \centering
    \includegraphics[width=\linewidth]{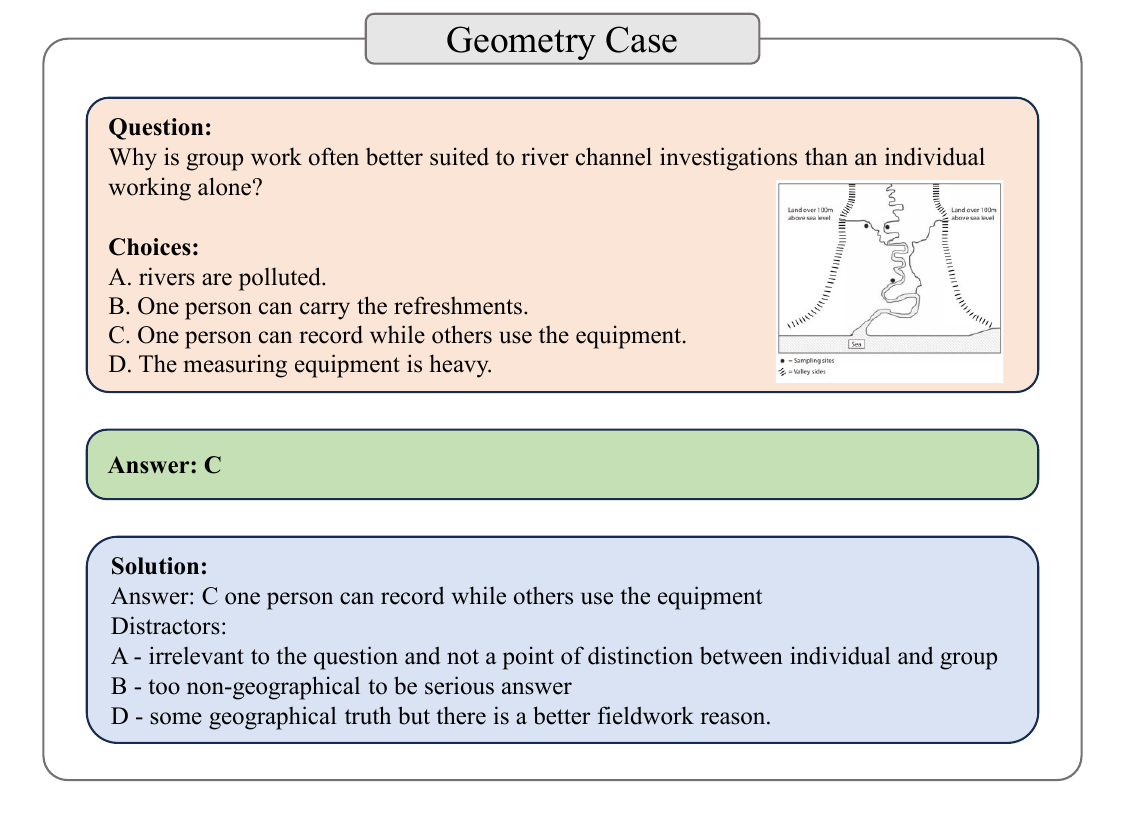}    
    \caption{An example question from geography.}
    \label{fig:geo-case}
\end{figure}
\clearpage

\begin{figure*}[p]
\vspace*{\fill}
\centering
\includegraphics[width=\linewidth]{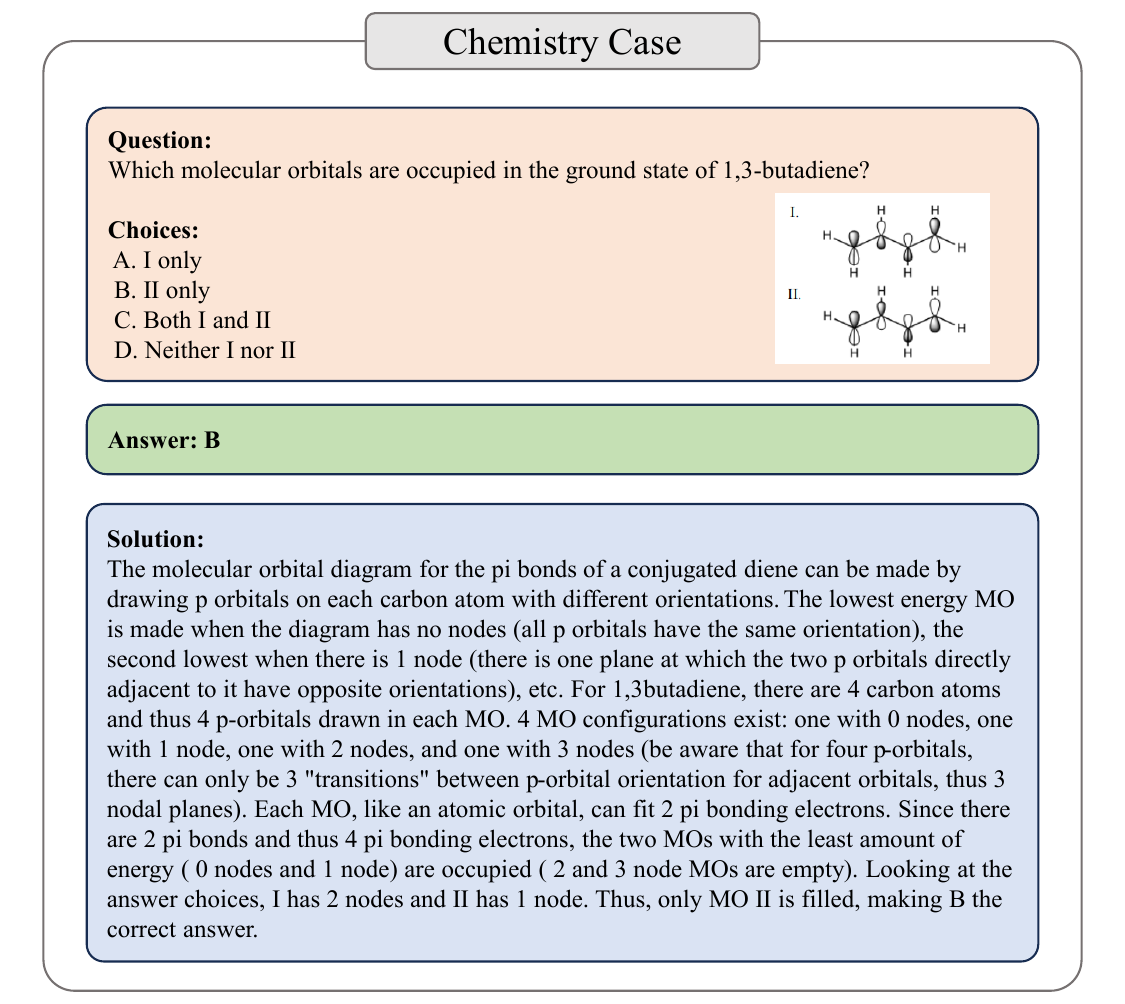}
\caption{An example question from chemistry.}
\label{fig:chem-case}
\vspace*{\fill}
\end{figure*}
\clearpage

\begin{figure*}[p]
\vspace*{\fill}
\centering
\includegraphics[width=\linewidth]{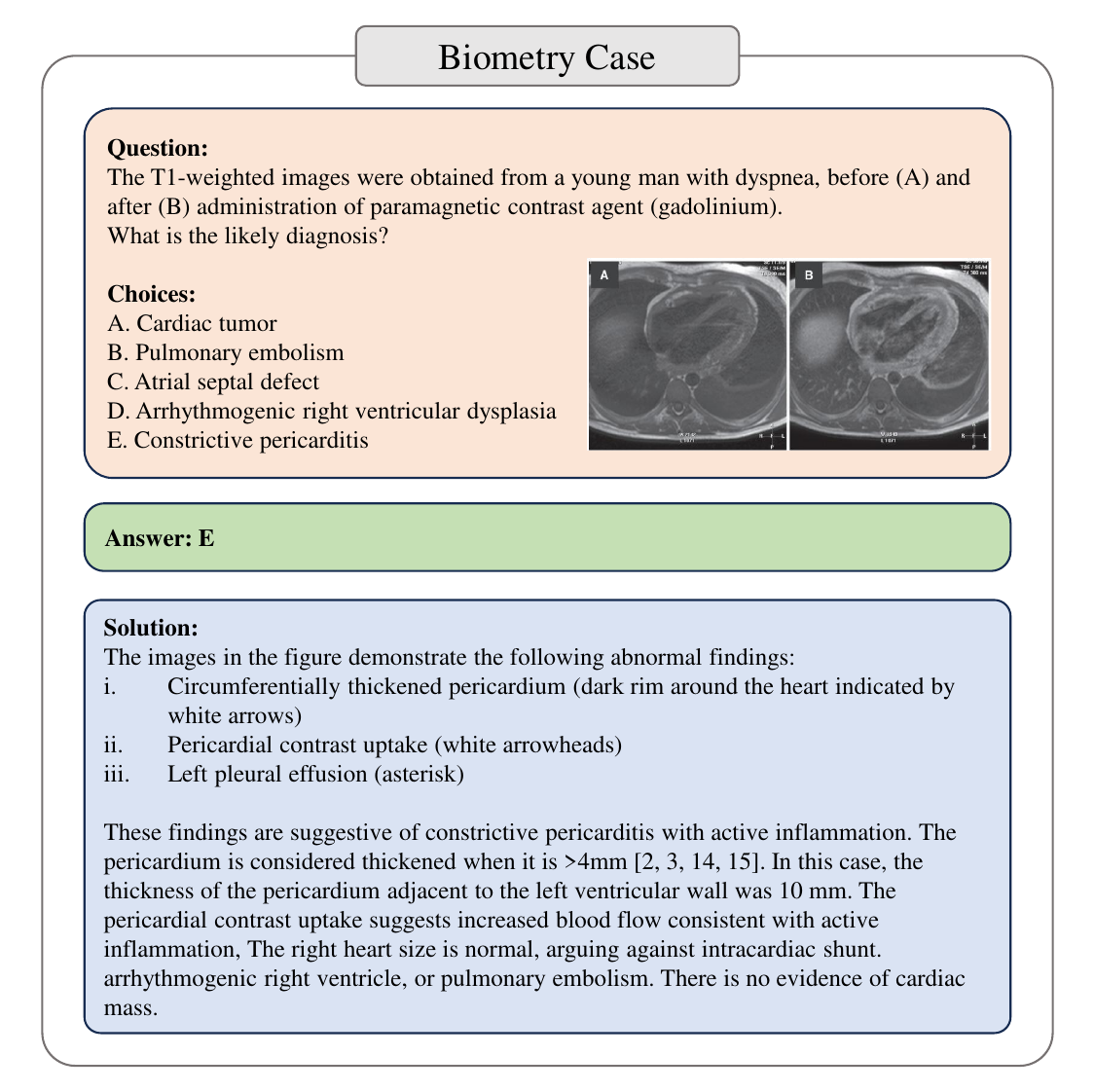}
\caption{An example question from biology.}
\label{fig:bio-case}
\vspace*{\fill}
\end{figure*}
\clearpage

\begin{figure*}[p]
\vspace*{\fill}
\centering
\includegraphics[width=\linewidth]{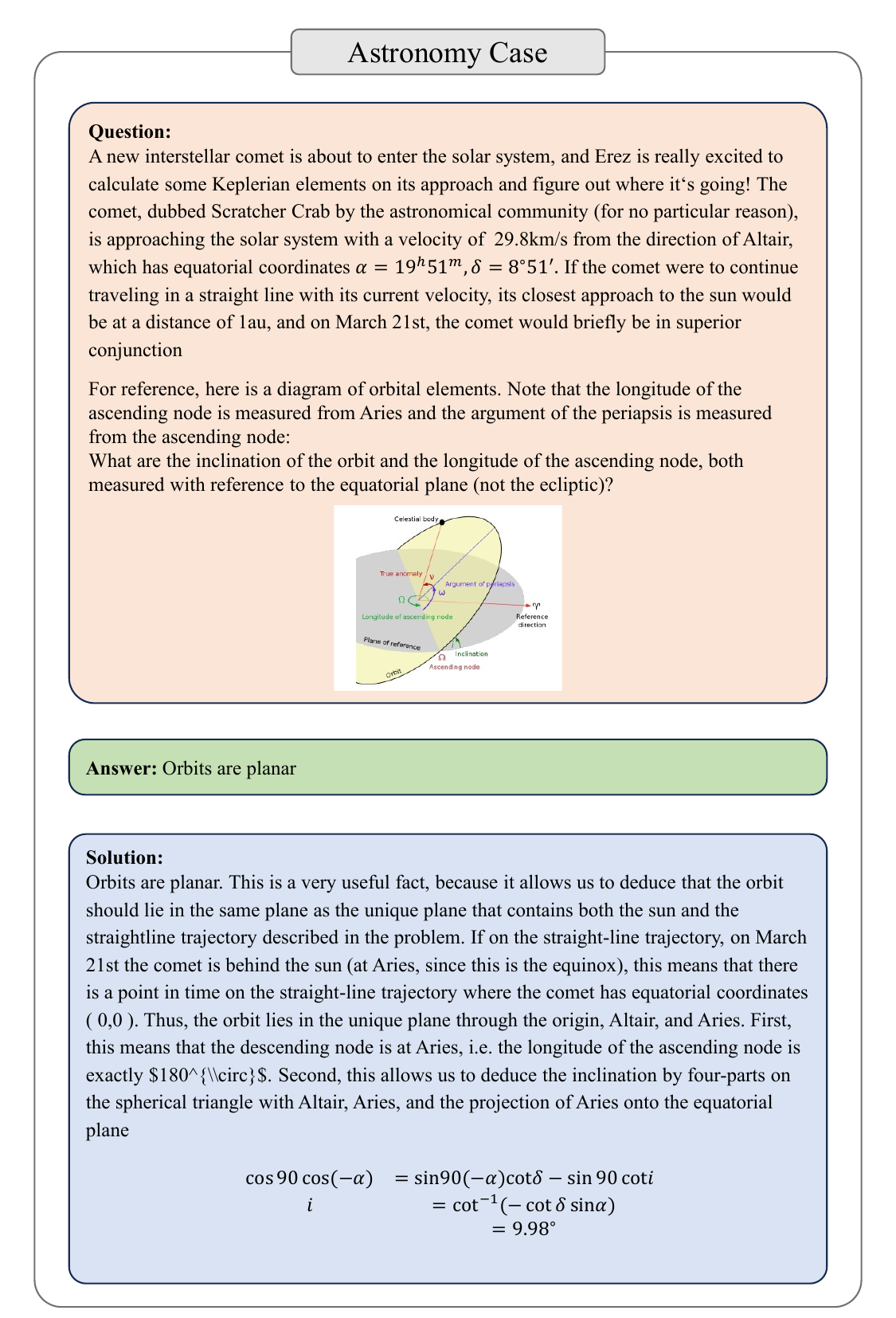}
\caption{An example question from astronomy.}
\label{fig:astronomy-case}
\vspace*{\fill}
\end{figure*}
\clearpage

\begin{figure*}[p]
\vspace*{\fill}
\centering
\includegraphics[width=\linewidth]{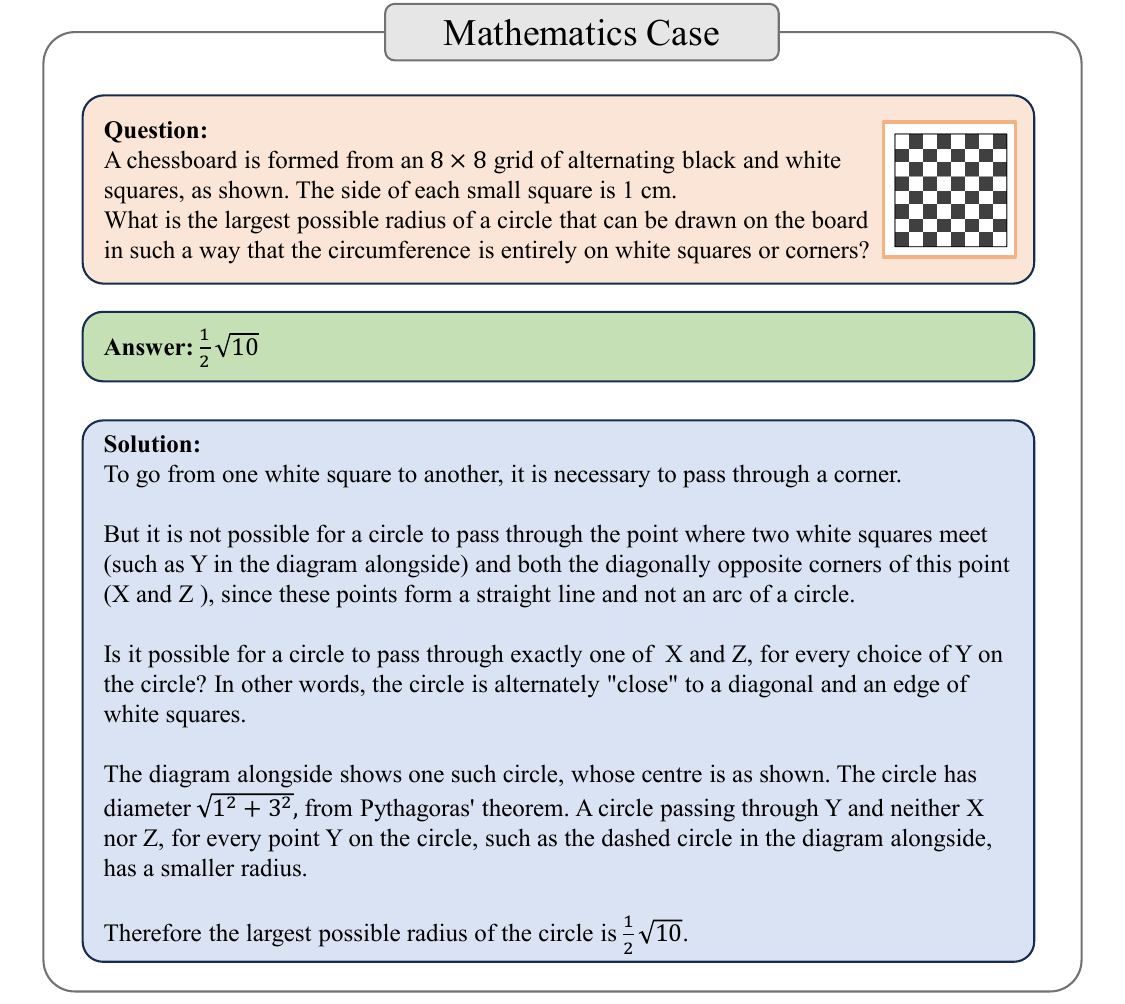}
\caption{An example question from mathematics.}
\label{fig:math-case}
\vspace*{\fill}
\end{figure*}
\clearpage

\begin{figure*}[p]
\vspace*{\fill}
\centering
\includegraphics[width=\linewidth]{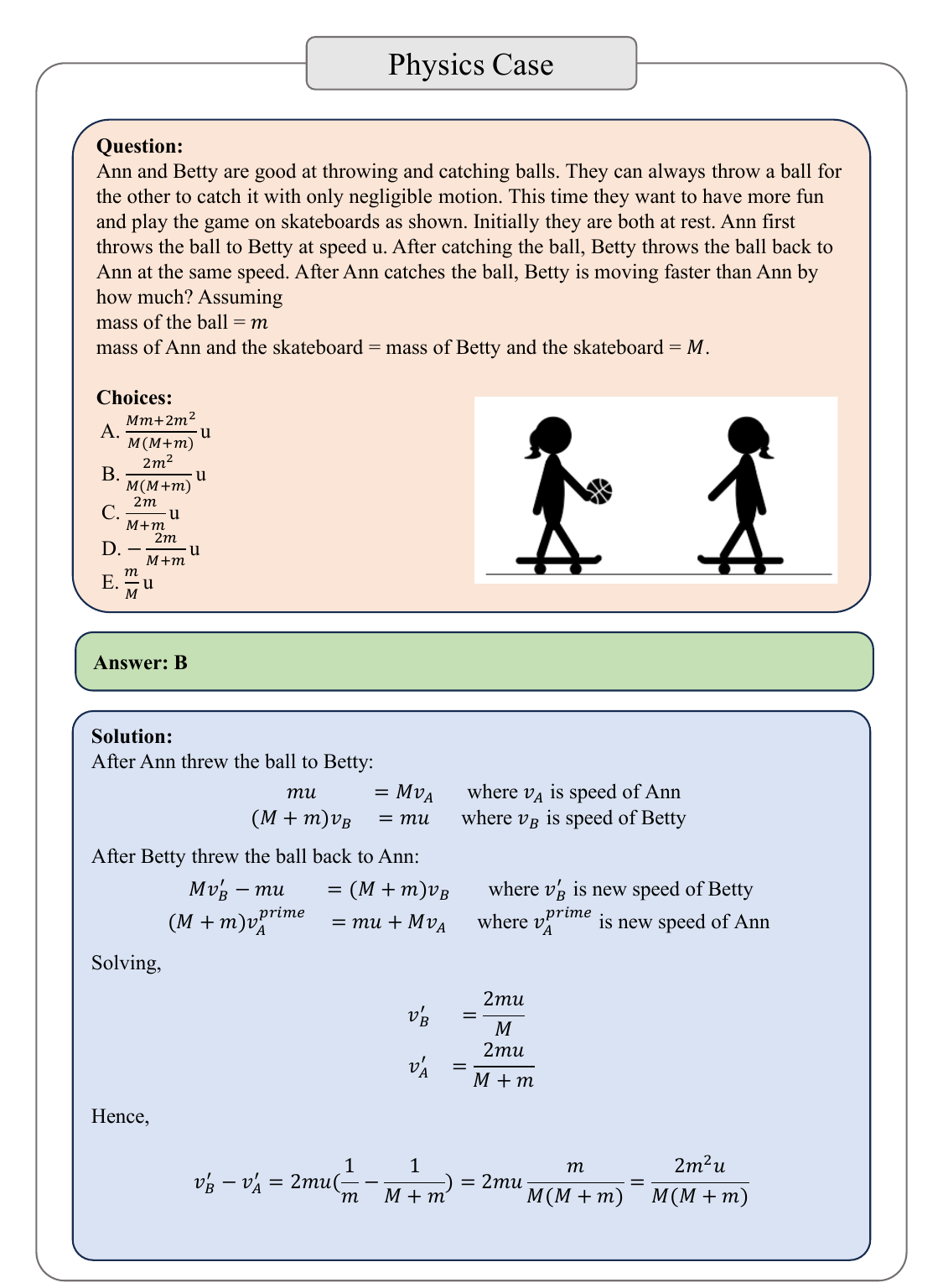}
\caption{An example question from physics.}
\label{fig:phy-case}
\vspace*{\fill}
\end{figure*}
\clearpage

\end{document}